\documentclass[10pt,twocolumn,letterpaper]{article}

\usepackage{cvpr}              

\usepackage{float}
\usepackage{graphicx}
\usepackage{stfloats}

\usepackage[dvipsnames]{xcolor}

\usepackage{duckuments}

\definecolor{cvprblue}{rgb}{0.21,0.49,0.74}
\usepackage[pagebackref,breaklinks,colorlinks,citecolor=cvprblue]{hyperref}

\def\paperID{7830} 
\def\confName{CVPR}
\def\confYear{2024}

\usepackage{ifthen}
\definecolor{MyGreen}{rgb}{0,0.5,0.2}
\definecolor{MyBlue}{rgb}{0,0.2,1.0}
\definecolor{MyRed}{rgb}{1,0.2,0.2}
\definecolor{MyPurple}{rgb}{0.5,0,1.0}
\definecolor{MyOrange}{rgb}{1.0,0.5,0}
\definecolor{MyBrown}{rgb}{0.65,0.35,0}
\definecolor{MyMagenta}{rgb}{1.,0.0,1.}
\definecolor{MyGrey}{rgb}{0.557,0.557,0.557}

\newcommand{\zwc}[1]{{\color{MyRed}{(ZW: #1)}}}

\newcommand{\showcomments}{1}  
\ifthenelse{\isundefined{\showcomments}}
{
\renewcommand{\zwc}[1]{}

}{}
\newcommand{\hidecontents}[1]{}

\begin{document}

\title{MagicScroll: Nontypical Aspect-Ratio Image Generation for Visual Storytelling via Multi-Layered Semantic-Aware Denoising
}

\author{Bingyuan Wang$^1$ \quad
Hengyu Meng$^3$ \quad
Zeyu Cai$^1$ \quad
Lanjiong Li$^1$ \quad \\
Yue Ma$^2$ \quad
Qifeng Chen$^2$ \quad
Zeyu Wang$^{1,2\dagger}$ 
\\
$^1$HKUST(GZ) \quad 
$^2$HKUST \quad
$^3$South China University of Technology\quad  \\
\url{https://magicscroll.github.io/}
}

\twocolumn[{
\maketitle
\begin{center}
    \captionsetup{type=figure}
    \includegraphics[width=\textwidth]{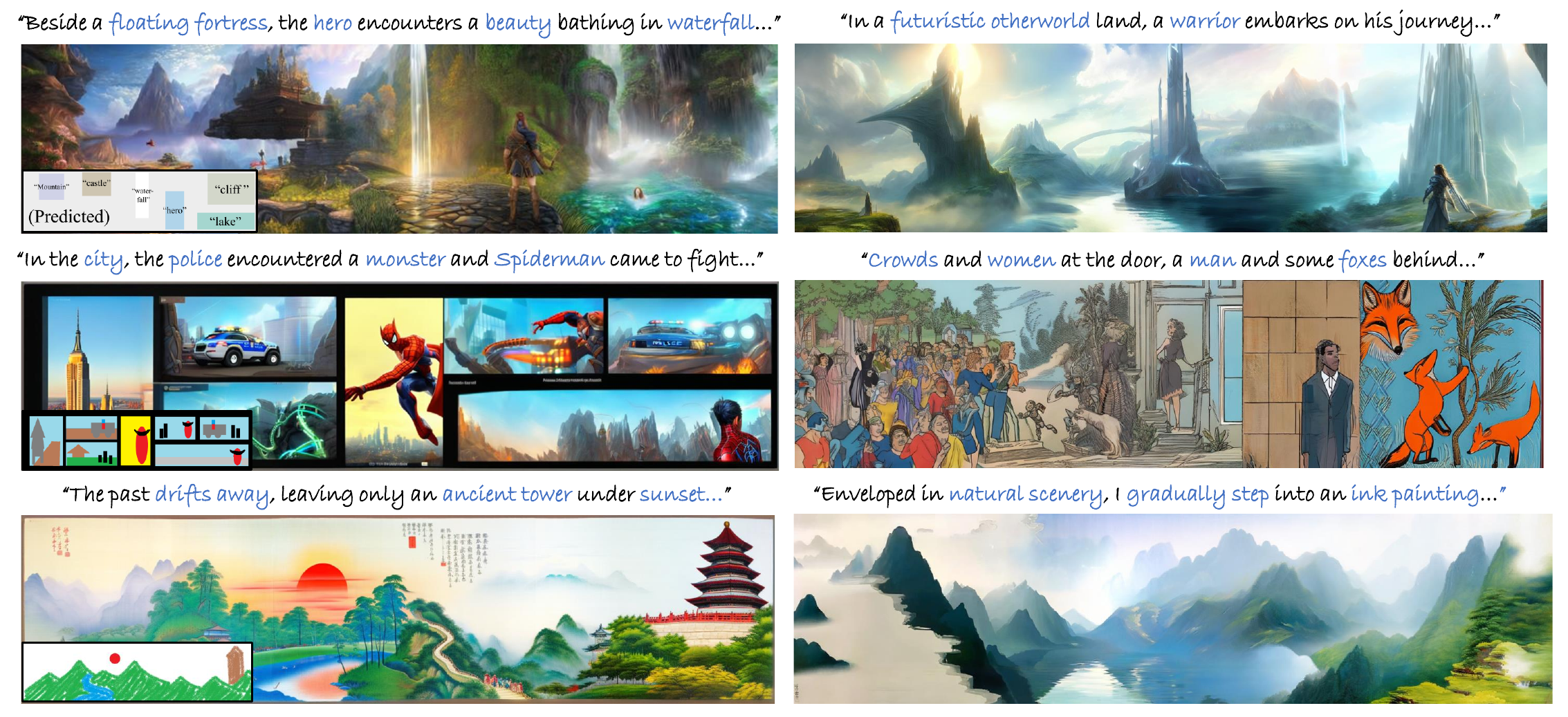}
    \captionof{figure}{\textbf{\textit{Example results generated by MagicScroll.}} Our framework is designed for generating coherent, controllable, and engaging nontypical aspect-ratio images from story texts. We support multi-layered, refined controls over style, content, and layout, with multiple conditions including predicted masks, reference images, and style concepts.}
    \label{fig:teaser}
\end{center}
}]

\newcommand\blfootnote[1]{%
  \begingroup
  \renewcommand\thefootnote{}\footnote{#1}%
  \addtocounter{footnote}{-1}%
  \endgroup
}

\blfootnote{$\dagger$ Corresponding author.}



\begin{abstract}

Visual storytelling often uses nontypical aspect-ratio images like scroll paintings, comic strips, and panoramas to create an expressive and compelling narrative.
While generative AI has achieved great success and shown the potential to reshape the creative industry, it remains a challenge to generate coherent and engaging content with arbitrary size and controllable style, concept, and layout, all of which are essential for visual storytelling.
To overcome the shortcomings of previous methods including repetitive content, style inconsistency, and lack of controllability, we propose MagicScroll, a multi-layered, progressive diffusion-based image generation framework with a novel semantic-aware denoising process.
The model enables fine-grained control over the generated image on object, scene, and background levels with text, image, and layout conditions.
We also establish the first benchmark for nontypical aspect-ratio image generation for visual storytelling including mediums like paintings, comics, and cinematic panoramas, with customized metrics for systematic evaluation.
Through comparative and ablation studies, MagicScroll showcases promising results in aligning with the narrative text, improving visual coherence, and engaging the audience.
We plan to release the code and benchmark in the hope of a better collaboration between AI researchers and creative practitioners involving visual storytelling.
\end{abstract}

%











\section{Introduction}
\label{sec:intro}



Recent advances in generative AI have achieved great success in visual content generation from text and have been widely applied to the creative industry. One creative application that has a large demand for AI-generated content is visual storytelling. Visual storytelling typically requires image sequences or videos that are consistent with both the narrative and its atmosphere through multiple coherent and stylized scenes. However, there is a lack of suitable methods for generating such visual content, especially with a high level of controllability, correspondence, coherence, and content richness.



As a classic visual storytelling form, nontypical aspect-ratio images allow for a more dynamic and engaging representation of the story's progression and emotion through spatially unfolding a series of objects and scenes described in the text. These aspects are commonly reflected in scroll paintings, comic strips, and cinematic panoramas, which leverage imagery in multiple scenes to construct a unified and stylized space for the story.
However, existing methods did not distinguish storytelling visuals from common images, thus lack conditional models for refined local control. Another challenge is a lack of consideration of smooth transition, especially between multiple generated results.



This paper contributes a framework for nontypical aspect-ratio image generation for visual storytelling with a benchmark for systematic evaluation. We first define the task and describe its challenges and applications. Based on state-of-the-art diffusion models, we propose a novel method that integrates GPT-based layout prediction, semantic-aware denoising, and multiple controllable modules. Given the original story, our method is capable of predicting scene and object masks and generating nontypical aspect-ratio images that align with both textual narrative and conditions for user control such as style and layout.

To address the issues of repetitive content and inconsistent style in existing methods, we design a novel multi-layered semantic-aware denoising process to inject controllability at object, scene, and background levels with style and layout conditions. To augment semantic consistency and enhance visual coherence, we incorporate latent blending and latent smoothing into each denoising step. Enabled by additional text-based and image-based style controls, our method supports multiple scenarios including painting, comic, and panorama without tuning, and can be easily customized for creative needs.





For this kind of image generation task, there are no dedicated datasets or metrics for evaluating generated nontypical aspect-ratio images for visual storytelling. To fill this gap, we introduce a comprehensive dataset and new metrics for content richness and coherence. We also conducted a series of comparative and ablation studies to demonstrate the effectiveness of our framework and each module, outperforming existing methods based on these metrics and subjective user ratings.


In summary, our main contributions are:
\begin{itemize}
\item 
A framework to generate nontypical aspect-ratio images from storytelling text with style and layout controls.


\item 
A novel semantic-aware denoising process to enhance the coherence and controllability in generated images.


\item 
A dedicated benchmark and metrics to evaluate nontypical aspect-ratio image generation for visual storytelling. 

\end{itemize}


\section{Related Work}
\label{sec:related}



\subsection{Controllable Generation with Diffusion Models} 
Diffusion models, such as DDPM~\cite{ho2020denoising} and DDIM~\cite{song2020denoising}, have proved effective in various image generation tasks~\cite{dhariwal2021diffusion}, especially with different forms of conditional control~\cite{choi2021ilvr}. Recent work integrates multiple concept, style, and layout conditions into the generative process, which provides insight into enhancing the controllablility in content generation for visual storytelling.


\noindent\textbf{Concept and Style Control.}
Among them, LoRA~\cite{hu2021lora} focuses on domain adaptation, Textual Inversion~\cite{gal2022image} represents user-provided concepts as pseudo-words, and DreamBooth~\cite{ruiz2023dreambooth} introduces a unique identifier for diffusion model personalization. Specifically, methods like StyleDrop~\cite{sohn2023styledrop}, StyleAdapter~\cite{wang2023styleadapter}, and Ip-Adapter~\cite{ye2023ip} target the modification of stylistic attributes. 
 


\noindent\textbf{Composition and Editing.} 
To deal with multiple concepts in image generation, there have been composable diffusion models~\cite{liu2022compositional} and models like Attend-and-Excite~\cite{chefer2023attend}, which refine cross-attention units for semantically accurate images. Other methods such as Prompt2Prompt~\cite{hertz2022prompt}, SDEdit~\cite{meng2021sdedit}, and Imagic~\cite{kawar2023imagic} focus on image editing through textual prompts. DragGAN~\cite{pan2023drag} also introduces an interactive approach for editing generated images.


\noindent\textbf{Spatial and Layout Control.} 
Research on controllable image generation often focuses on adjusting attention layers to control spatial relationships. Representative methods include GLIGEN~\cite{li2023gligen}, DirectedDiffusion~\cite{ma2023directed}, and LayoutDiffusion~\cite{zheng2023layoutdiffusion}. Additionally, LayoutDM~\cite{inoue2023layoutdm} conducts diffusion on a discrete state-space with layout constraints, LayoutLLM~\cite{qu2023layoutllm} employs a coarse-to-fine paradigm, and ZestGuide~\cite{couairon2023zero} introduces a mechanism for conditioning on free-form text. ControlNet~\cite{zhang2023adding} and Uni-ControlNet~\cite{zhao2023uni} propose unique convolution layers for robust, efficient training and task-specific condition learning. 

While these methods support specific types of control, they have not been effectively combined to achieve complex visual narratives, which is a crucial need in storytelling scenarios. Meanwhile, few of them support refined control over large images, especially with nontypical aspect ratios.

\subsection{Story Visualization}
Story visualization is an increasingly popular task that has emerged in recent years following the rise of generative AI. Methods like StoryGAN~\cite{StoryGAN} and its subsequent works~\cite{maharana2022storydall,makeastory,sv, ma2023magicstick}
aim to produce a sequence of images that are semantically aligned with a provided series of text. In contrast to the focus of text-to-video methods ~\cite{singer2022make,ma2022visual, weng2019photo,wang20223d, ma2023follow, ma2022simvtp}, this approach emphasizes less the coherence between generated images per frame but more on the overall consistency across dynamic scenes and characters. Despite these achievements, story visualization with image sequences still suffers from issues including low visual quality, limited controllability, and lack of content richness. 


Meanwhile, nontypical aspect-ratio images have the potential to represent rich visual storytelling content as shown in multiple panorama generation tasks. For example, PanoGen~\cite{li2023panogen} employs image outpainting to generate images with nontypical aspect ratios. MultiDiffusion~\cite{bar2023multidiffusion} strengthens the continuity of generated images by optimizing multiple diffusion generation processes, binding them with a set of shared parameters or constraints. ScaleCrafter~\cite{he2023scalecrafter} combines dispersed convolutions and unguided classifiers to produce images with arbitrary aspect ratios. However, these methods typically lack the controllability required in visual storytelling, as the generated content needs to be consistent with the narrative and convey emotions through multiple coherent and stylized scenes. To address these issues, we introduce a new framework to generate nontypical aspect-ratio images with coherent and engaging content, aiming for better visual storytelling.

\section{Method}
\label{sec:method}
\subsection{Overview}
\begin{figure*}[h]
    \centering
    \includegraphics[width=\linewidth]{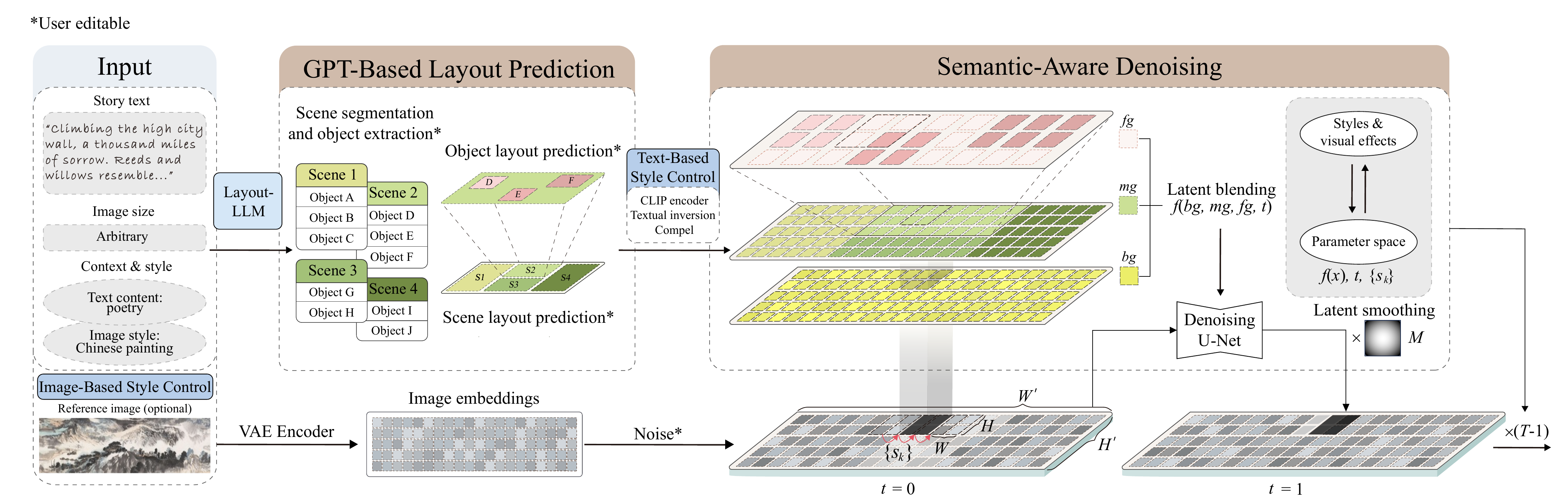}
    \caption{\textbf{\textit{Pipeline of our method.}} MagicScroll mainly consists of GPT-based layout prediction, semantic-aware denoising, as well as text-based and image-based style control. The layout prediction module extracts bounding boxes of scenes and objects from the story text. The style control modules determine the stylistic attributes via encoding the text prompt and reference image. The semantic-aware denoising module includes latent blending and smoothing to generate controllable, coherent, and engaging nontypical aspect-ratio images.  
    }
    \label{fig:pipeline}
\end{figure*}


Figure~\ref{fig:pipeline} shows the pipeline of our method, which is mainly comprised of three parts: GPT-based layout prediction, semantic-aware denoising, and text-based and image-based style control.
In a typical story-to-scroll generation process, the textual inputs undergo scene segmentation and object extraction, before being converted to a series of latent codes with specific concepts, objects, and styles. From an initial reference image or random noise, the model uses a semantic-aware denoising process including temporal latent blending and spatial latent smoothing, gradually morphing it into a coherent and engaging image while maintaining the target style, content, and layout.

\subsection{GPT-Based Layout Prediction}

A challenge in text-to-image generation tasks is to understand and predict reasonable spatial relationships. Therefore, we leverage GPT's prior knowledge of understanding semantic and spatial relationships for layout prediction.

First, we separate a sequence of story text into distinct scenes (similar to a movie storyboard) and extract key objects belonging to each scene, which lays a foundation for the following generative process. We realize the function by prompting a GPT-3.5~\cite{openai_model_index} model with requirements and output data format. However, our experiments showed that the results of contextual and affective understanding varied greatly across different people and GPT trials. So we focus on the key scenes and objects within textual input and only infer text genre (e.g., heroic story) and image styles (e.g., comic strip) to facilitate the generative process.



Next, inspired by LayoutLLM~\cite{qu2023layoutllm}, we use GPT to produce possible bounding boxes of both scenes and objects to accomplish open-domain layout generation, aiming to effectively convey the semantic details present in the input text across various hierarchical levels. 
To enhance the capability of the large language model in accurately predicting layouts, we employ visual instruction tuning~\cite{liu2023visual}, a process of providing comprehensive instructions with sample answers to facilitate a better understanding of the task. As demonstrated by our results (Figure~\ref{fig:ablation}), the layout predictor can significantly enhance the hierarchy and controllability of semantic structures in the generated image, a key aspect of visual storytelling.

\subsection{Semantic-Aware Denoising}
\label{subsec:generative model}

The Latent Diffusion Model~\cite{rombach2022high} proposes an image generation method with a diffusion process in the latent space $\mathcal{Z}$. Specifically, it uses an encoder $\mathcal{E}$ to encode a pixel image $x$ into low-resolution latents $z=\mathcal{E}(x)$. During training, the model optimizes a denoising U-Net $\epsilon_\theta$ by eliminating artificial noise, conditioned on both text prompt embedding $y$ and current image sample $z_t$, which is a noisy sample of $z_0$ at step $t\in[0, T]$):
\begin{equation}
\min _\theta E_{z_0, \epsilon \sim N(0, I), t}\left\|\epsilon-\epsilon_\theta\left(z_t, t, y\right)\right\|_2^2.
\label{equation: denosing}
\end{equation}
After training, it is capable of converting random noise $\epsilon$ to an image sample $z$ by the learned denoising process.


\noindent\textbf{Latent Smoothing.}
Since coherence and story integrity are key to generating visual storytelling content, we design a latent smoothing method to better concatenate different masks at the same level (e.g., adjacent scenes). Our proposed method is a combination of sliding window and edge matrix to improve spatial coherence.

We adopt a similar method as MultiDiffusion~\cite{bar2023multidiffusion} to iterate over the whole image with a series of intersecting sliding windows $F_i(J), i \in [1, N]$, where $F$ is a mapping from non-typical aspect-ratio image $J\in R^{H' \times W' \times C}$ to the standard diffusion paths $I\in R^{H \times W \times C}$ within each sliding window. As a result, the new denoising process $\Omega$ at step $t$ is achieved by the following optimization problem:
\begin{equation}
    \Omega(J_t, y)=\underset{J\in R^{H' \times W' \times C}}{\arg\min} \sum_{i=1}^{N} \left\|F_i(J)-\Phi(F_i(J_t), y)\right\|_2^2,
    \label{inference}
\end{equation}
where $\Phi$ denotes a standard denoising step in $R^{H \times W \times C}$. We found out that different strides $s$ of the sliding window not only affect the number of sliding windows $N$:
\begin{equation}
    N=(\frac{H'-H}{s}+1)(\frac{W'-W}{s}+1),
\end{equation}
but also reflect different levels of detail in semantics, thus we select a series of coarse-to-fine strides $\{s_k\}$. Finally, each denoised window is multiplied by an edge matrix $M\in R^{H \times W \times C}$ using the Hadamard product:
\begin{equation}
    F_i'(J) = F_i(J) \odot M.
\end{equation}


This modification adjusts latent weights based on their distance from the edge. Our experiments discover that while this operation can significantly enhance smooth transitions, edge matrices using linear, cosine, or Gaussian functions may achieve the best results in different scenarios.

\noindent\textbf{Latent Blending.}
Another aspect of coherence is in the transition between different layers inclduing foreground objects, midground scenes, and background styles. To solve the problem, we introduce a latent blending method to map the multi-level latents to a time series for content-style blending and visual effect control.

To preserve spatial information, we use latents at three different levels $bg, mg, fg\in R^{H' \times W' \times C \times L}$, where $L$ is the latent dimension. For each timestep $t$, we calculated its specific prompt embedding to update the global $y$ in Eq. \ref{inference}:

\begin{equation}
    y(i,t) = f(bg, mg, fg, t)
\end{equation}

\begin{figure*}[h]
\centering

\includegraphics[width=\linewidth]{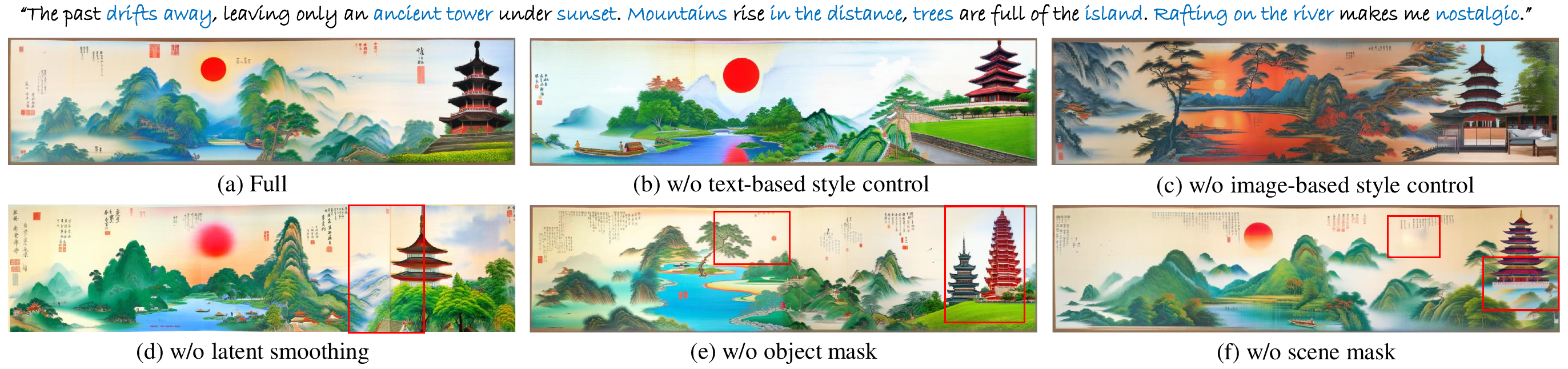}

    \caption{\textbf{\textit{Example results of the ablation study.}} Red boxes indicate artifacts. Among the modules, (b) text-based style control and (c) image-based style control mainly affects the overall style and color. (d) Latent smoothing can improve visual coherence. (e) Object maks and (f) scene masks not only affect image layout but also significantly influence the consistency between foreground and midground.}
    \label{fig:ablation}
\end{figure*}

Based on controlled experiments, we discovered that conditioning on objects and scenes in the generation process has different impacts on the outcome. While both enhancing the narrative aspect of visual storytelling, the former focuses more on concrete semantics, and the latter contributes more to coherent visuals, reflecting a compromise between semantic details and artistic styles in the creative process. Based on previous two-layer generation methods~\cite{li2023gligen}, we use a latent blending function $f$ that is foreground for the first 15\% and midground for the remaining 85\% denoising steps, while alternating between foreground/midground and background latents, which empirically yields better results as shown in Figure~\ref{fig:ablation}.

\subsection{Style Control Modules}
 
For visual storytelling with nontypical aspect-ratio images, a major challenge is generating coherent and stylized content. To address this, we design style control modules based on both text and image encoding. 

\noindent\textbf{Text-Based Style Control.}
For text prompt encoding, we integrate a Textual Inversion~\cite{gal2022image} module and a Compel~\cite{compel} module to the basic CLIP~\cite{radford2021learning} text encoder, to better incorporate and enhance different content and style concepts.
Following previous research~\cite{deng2023nerdi, ao2023gesturediffuclip, wu2023not} on separating high-level concepts and low-level visuals, we adopt both textual inversion and image captioning, to better learn from the images in our dataset. Specifically, we first derive scene and object labels from our reference datasets with an OFA~\cite{wang2022ofa} model, and then train Textual Inversion models for styles and concepts separately.

To better control target objects to appear at specified locations, we propose an original method to combine textual inversion, mask, and Compel on different image layers. Specifically, users may use operators to enhance or reduce certain content or style in the text prompt. If not specified, our model will also strengthen the mask content in the scene-level text prompt to facilitate generating target objects. Our results show that text-based style control is effective in helping users generate desired objects, introduce specific content and styles, and exert more precise control.

\noindent\textbf{Image-Based Style Control.}
As one of the user-editable modules, we allow users to input a reference image of different types (e.g., segmentation map, sketch, and contour) in an arbitrary aspect ratio. Inspired by SDEdit~\cite{meng2021sdedit}, we propose a process that extends the color and layout of the reference image to our output. Specifically, we first pass it through a variational autoencoder~\cite{kingma2013auto} to obtain an initial latent representation and add Gaussian noise to obtain a noisy latent, simulating a reverse Stochastic Differential Equation (SDE).

Figure~\ref{fig:teaser} shows some results generated with a reference image. Combined with previous modules, ours not only mimics the styles of specific genres or painters but also achieves coherent style transitions within the same image while maintaining the original layout and colors. This enhances the controllability of our method, making the results more consistent with the narrative and aligning better with user expectations.






\section{Evaluation}
\label{sec:evaluation}

In this section, we provide a benchmark covering three typical scenarios for visual storytelling: painting, comic, and cinematic panorama. We also designed two metrics for evaluating the degree of content richness and diversity in a single image. Finally, we conducted a comparative study and an ablation study to evaluate the performance of MagicScroll and its each module.

\subsection{Benchmark}
\label{subsec:benchmark}
\noindent\textbf{Datasets.} Our benchmark includes a self-curated painting dataset and two refined datasets of comics and cinematic panoramas. All collected data have been filtered and post-processed to better evaluate the quality of visual storytelling. Below we provide their detailed descriptions:

\begin{itemize}
    \item \textbf{Painting} (Traditional Chinese Landscape Paintings). This dataset contains 130 pairs of ancient Chinese poetry and corresponding ancient Chinese paintings. The data are collected from The Palace Museum website~\cite{minghuaji}. 
    \item \textbf{Comic} (Multiple Comic Strips). This dataset contains 100 comic images with masks for panels, balloons, and text. The data are downloaded from eBDtheque~\cite{ebdtheque}.
    \item \textbf{Cinematic Panorama} (Movie Scripts). The dataset contains more than 6000 rows of annotated movie script data. The data are collected from Movie Scripts Corpus~\cite{movie_scripts_corpus}.
\end{itemize}

\noindent\textbf{Metrics.} Based on the needs in visual storytelling and challenges in nontypical aspect-ratio image generation, we mainly focus on controllability, coherence, and narrativity in the evaluation process. To better quantify the outcomes, we first select some indicators for diffusion-based text-to-image tasks from different perspectives:
\begin{itemize}
    \item \textbf{Text-Image Similarity:} The similarity between input text and output image (e.g., CLIP Score, CLIP Directional Similarity, CLIP R-Precision).
    \item \textbf{Image Quality:} Calculated based on generated results and ground truths, typically concerning fidelity and diversity, e.g., Fréchet Inception Distance (FID) for large datasets, Inception Score (IS) for small datasets, PSNR for reconstruction.
    \item \textbf{Image Aesthetics:} Calculated from generated image and pre-trained priors, to evaluate from an artistic perspective, e.g., LAION-AI aesthetic predictor.
\end{itemize}


Nevertheless, there is a lack of metrics specifically designed to reflect the storytelling aspect in nontypical aspect-ratio images. For example, the local correspondence between the input story and output scroll cannot be measured by a global CLIP score, as certain semantics may be diluted in the text encoding process. Meanwhile, some other metrics are calculated globally or set-wise, such as content richness (by IS) and semantic coherence (by self-correlation or color gradient), which are also not ideal for reflecting local information. In light of this, we propose two metrics below:
\begin{itemize}
    \item \textbf{Local-Global Image Scores (LGIS):} 
    Based on different scenes in a visual storytelling image, we calculate their respective CLIP similarity to the whole image and average them to get the result.
    \item \textbf{Global Embedding Variance (GEV):} We derive the embeddings of the whole image and different scenes and calculate their variance to represent semantic dispersion. 
\end{itemize}

These two metrics reflect the content richness and diversity of image latents from a local-global perspective. Besides, we also focus on the transition smoothness between different scenes and measure their LAION-AI aesthetic (\textbf{Edge Aesthetics, EA}) as an indicator. To sum up, our final metrics are CLIP-score, CLIP similarity to ground truth, and newly proposed LGIS, GEV, and EA.







\subsection{Comparisons}

\noindent\textbf{Experimental Setup.}
While there is a lack of dedicated methods that generate nontypical aspect-ratio images for visual storytelling, some previous methods focused on panorama generation, image conditioning, or layout control. We selected a few representative models and also included a baseline model for comparison.

Specfically, we collected the public versions of MultiDiffusion~\cite{bar2023multidiffusion}, ScaleCrafter~\cite{he2023scalecrafter}, ControlNet~\cite{zhang2023adding}, GLIGEN~\cite{li2023gligen}, and Stable Diffusion~\cite{rombach2022high}. To minimize the variance brought by different pre-trained models, we used version 2.1 for all diffusion models except those with only one adapted version. To better align the capabilities of each model, we standardized the output of our experiments to 2048$\times$512 images, generated from story texts and a common hand-drawn image as reference.

\begin{figure*}[h]
\centering
\includegraphics[width=\textwidth]{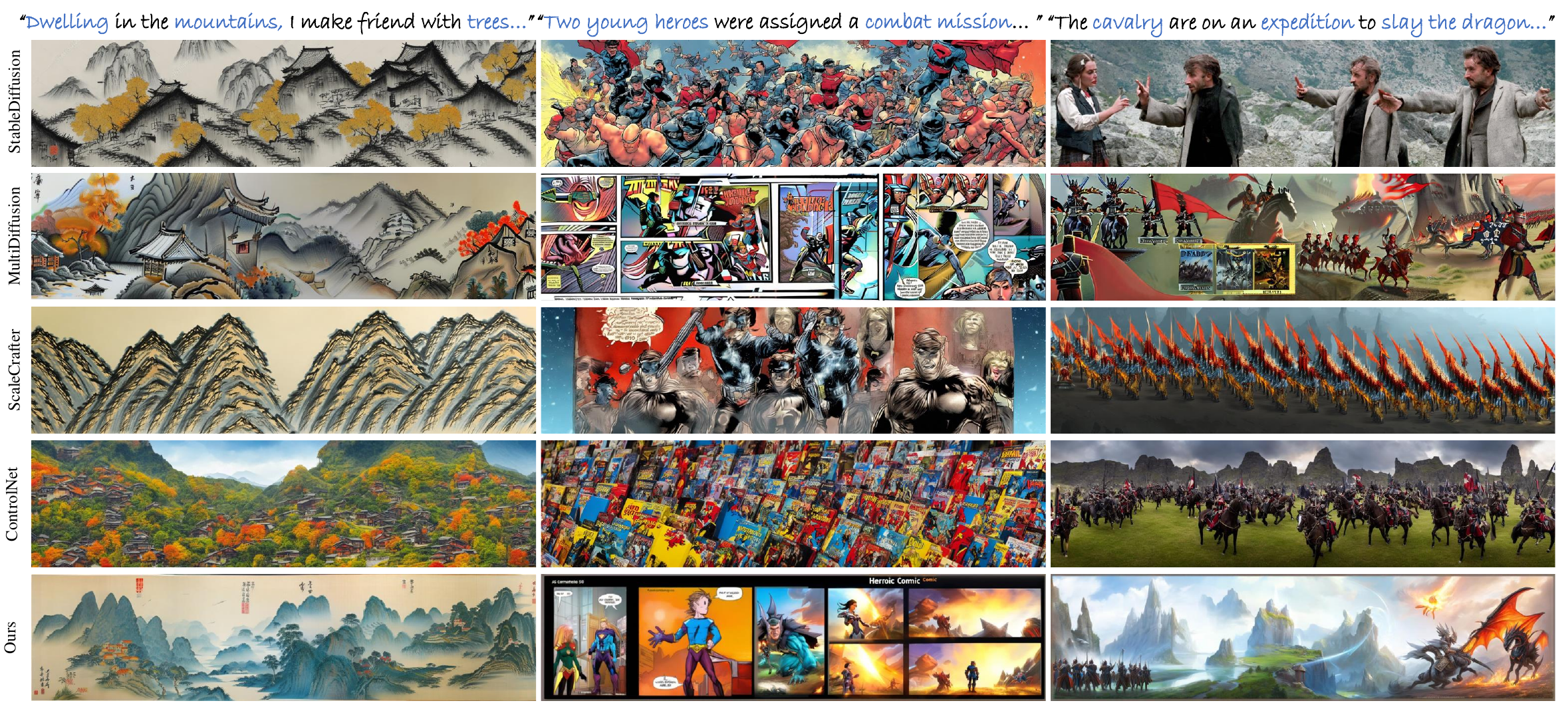}
    \caption{\textbf{\textit{Qualitative comparison of our methods with other baselines.}} The results demonstrate the high versatility and core advantages of MagicScroll. (a) Style mimicry in specific historical contexts. (b) Semantic layout planning and control. (c) Content richness and diversity. Among different scenarios, our framework can achieve more effective and engaging visual storytelling.}
    \label{fig:comparative}
\end{figure*}

\noindent\textbf{Qualitative Results.}
Some example results are shown in Figure~\ref{fig:comparative}. As expected, previous methods mainly suffer from a lack of content diversity, constantly generating repetitive content in different regions of the image. However, by GPT-based layout prediction, our method can retain the inherent chronology information of story text, as well as emphasize the key objects and scenes.

On the other hand, many critical needs in visual storytelling cannot be easily achieved with previous models, including object positioning, layout formatting, and color and style transition. In contrast, by integrating text-based and image-based style control into our novel denoising process, we can easily introduce specific layouts, concepts, and styles, ultimately achieving better narrative effects. Figure~\ref{fig:gallery} shows more results.

\noindent\textbf{Quantitative Results.}
Tables~\ref{tab:comparison} and~\ref{tab:gligen} report the automatic metrics aggregated for each model over the three different datasets. For the models that support generating nontypical aspect-ratio images, we focus on their content richness and semantic correspondence to ground truth. We also compare MagicScroll to GLIGEN, a previous state-of-the-art method for layout control, in both transition smoothness and text-image correspondence on the entire image and separate scenes.

The statistics in Table~\ref{tab:comparison} show that our model performs the best in both three aspects. In terms of similarity to the ground truth, results generated by Stable Diffusion are second to ours. This indicates that despite expanding the image size, other methods did not contribute significantly to the alignment between story texts and visuals. In contrast, our framework can better grasp such human perceptual patterns which may not be well reflected in the global CLIP Score, but are very important in visual storytelling.

On the other hand, from the LGIS and GEV metrics, our method is a top performer in generating more rich and diverse content. This also better caters to the need for visual storytelling scenarios to incorporate more information into a single image. Interestingly, our results suggest that in generating nontypical aspect-ratio images, the two objectives of ``closer to the ground truth'' and ``containing more information'' are not conflicting; rather, focusing on one of them may help achieve the other.

\noindent\textbf{User Evaluation.}
It is commonly acknowledged that user studies are essential in evaluating AI-generated content~\cite{wang2023imagen}. To evaluate visual outcomes produced by MagicScroll, we focused on three important dimensions: text-image consistency, visual coherence, and storytelling engagement. Specifically, users were asked to evaluate which of the generated images 1) \textit{is visually more coherent}, 2) \textit{better matches the story}, and 3) \textit{achieves more effective, expressive, and engaging storytelling}.

\begin{figure}[h]
    \centering
    \includegraphics[width=\columnwidth]{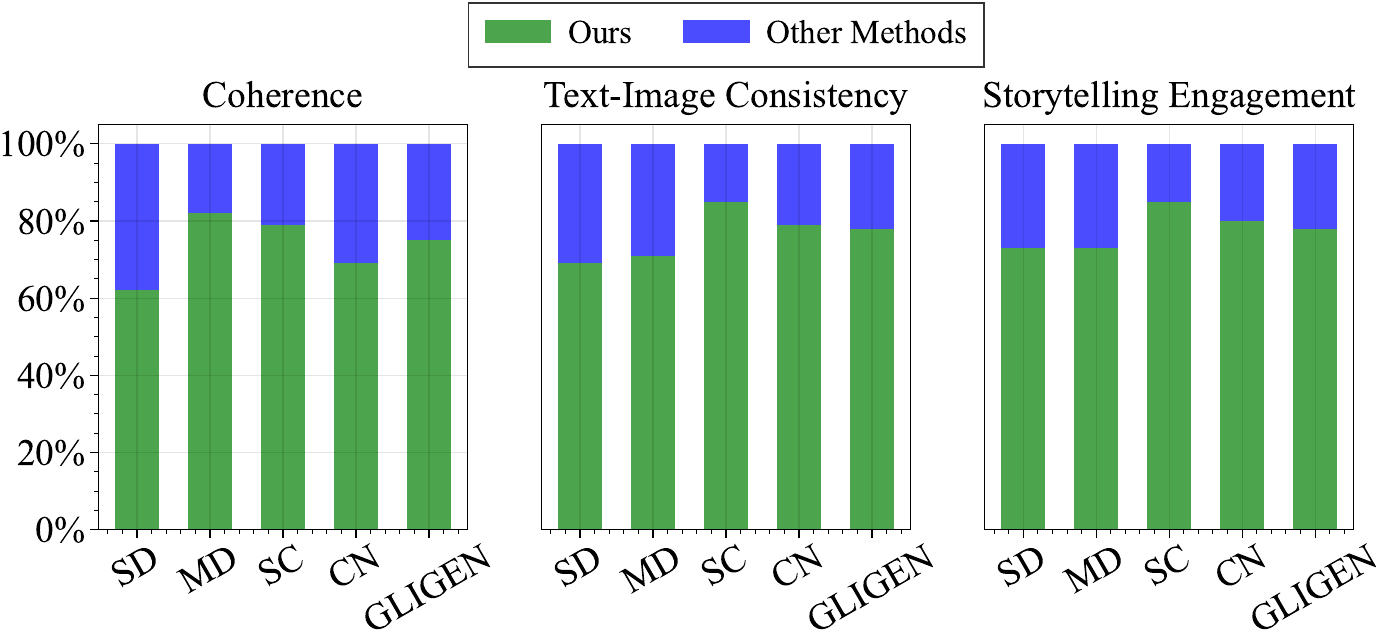}
    \caption{\textit{\textbf{User study results.}} We compared images generated from different models on text-image consistency, coherence, and storytelling engagement. The results show that our framework outperforms other alternative methods in all three aspects.}
    \label{fig:user_study}
\end{figure}

Figure~\ref{fig:user_study} shows 31 users' preferences for 25 nontypical aspect-ratio images generated by MagicScroll or other five methods. We collected 15 ratings for each method we compared against. The results demonstrate that our method not only supports better story-image alignment and visual coherence but also excels in immersing audiences in the narrative. From user feedback, the highest-rated features of MagicScroll include support for different scenarios, fast prototyping, intuitive editing, and refined control.



\subsection{Ablation Study}
\label{subsec:comparative study}

\noindent\textbf{Experimental Setup.}
We conducted an ablation study on the three main modules of our whole pipeline and two different inputs in the semantic-aware denoising process. We follow a similar experimental setup to the comparative study. The experimental groups are: w/o text-based style control, w/o image-based style control, w/o latent smoothing, w/o object mask, and w/o scene mask. 

\noindent\textbf{Results.}
Table~\ref{tab:ablation} shows the results of our ablation study. The results demonstrate the effectiveness of each module and their contributions to the final outcome from different perspectives. For example, text-based style control, scene masks, and object masks contribute to the content richness and similarity to ground truth. Meanwhile, each of the five modules is effective in enhancing the smooth transition between different scenes as well as text-to-image alignment.

\subsection{Discussion}

\begin{table}[]
    \centering
    \small
    \begin{tabular}{c|c|c|c}
    \toprule
         & CSGT↑ & LGIS↓ & GEV↑ \\
    \midrule
     MultiDiffusion   & 0.76 & 0.57 & 0.051 \\
     ScaleCrafter   &  0.74  & 0.62   & 0.047  \\
     ControlNet   &  0.67 &  0.57  &  0.053 \\
     StableDiffusion   &  0.78 &  0.60  & 0.049  \\
     \textbf{Ours} &  \textbf{0.80} &  \textbf{0.53}  & \textbf{0.057}  \\
    \bottomrule
    \end{tabular}
    \caption{\textbf{\textit{Results of end-to-end evaluation.}} This table provides comparative statistics of content richness and fidelity on the painting dataset. CSGT: CLIP Similarity to Ground Truth.}
    \label{tab:comparison}
\end{table}

\begin{table}[]
    \centering
    \small
    \begin{tabular}{c|c|c|c}
    \toprule
         & EA↑ & Local CLIP↑ & Global CLIP↑ \\
    \midrule
     GLIGEN   &  2.42 &  20.35  &  26.35 \\
     \textbf{Ours} & \textbf{2.72}  &  \textbf{20.40}  &  \textbf{26.69} \\
    \bottomrule
    \end{tabular}
    \caption{\textbf{\textit{Comparison with GLIGEN.}} Both GLIGEN and our method support layout control. Images generated by our method have better coherence and higher text-image similarity.}
    \label{tab:gligen}
\end{table}

\begin{table}[]
    \centering
    \small
    \begin{tabular}{c|c|c|c|c|c}
    \toprule
         & CSGT↑ & LGIS↓ & GEV↑ & EA↑ & CLIP↑ \\
    \midrule
      -TBSC &  0.73 &  0.59  &  0.052  & 2.54   & 26.62  \\
      -SM &  0.76 &  0.54  &  0.056  &  2.41  &  26.24 \\
      -OM &  0.78 &  \textbf{0.53} &  0.056  &  2.54  &  26.38  \\
      -LS &  0.79 &  \textbf{0.53}  &  0.056  &  2.55 & 26.57  \\
      -IBSC & \textbf{0.80}  &  \textbf{0.53}  &  \textbf{0.057}  &   2.67  & 26.59  \\
       Full &  \textbf{0.80} & \textbf{0.53} & \textbf{0.057} & \textbf{2.72} & \textbf{26.69}  \\
    \bottomrule
    \end{tabular}
    \caption{\textbf{\textit{Results of ablation study.}} This table lists statistics in the ablation study on the painting dataset. CLIP: Global CLIP score, TBSC: text-based style control, IBSC: image-based style control, SM: scene mask, OM: object mask, LS: latent smoothing.}
    \vspace{0.5cm}
    \label{tab:ablation}
\end{table}


\begin{figure*}[h]
    \centering
    \includegraphics[width=0.9\textwidth]{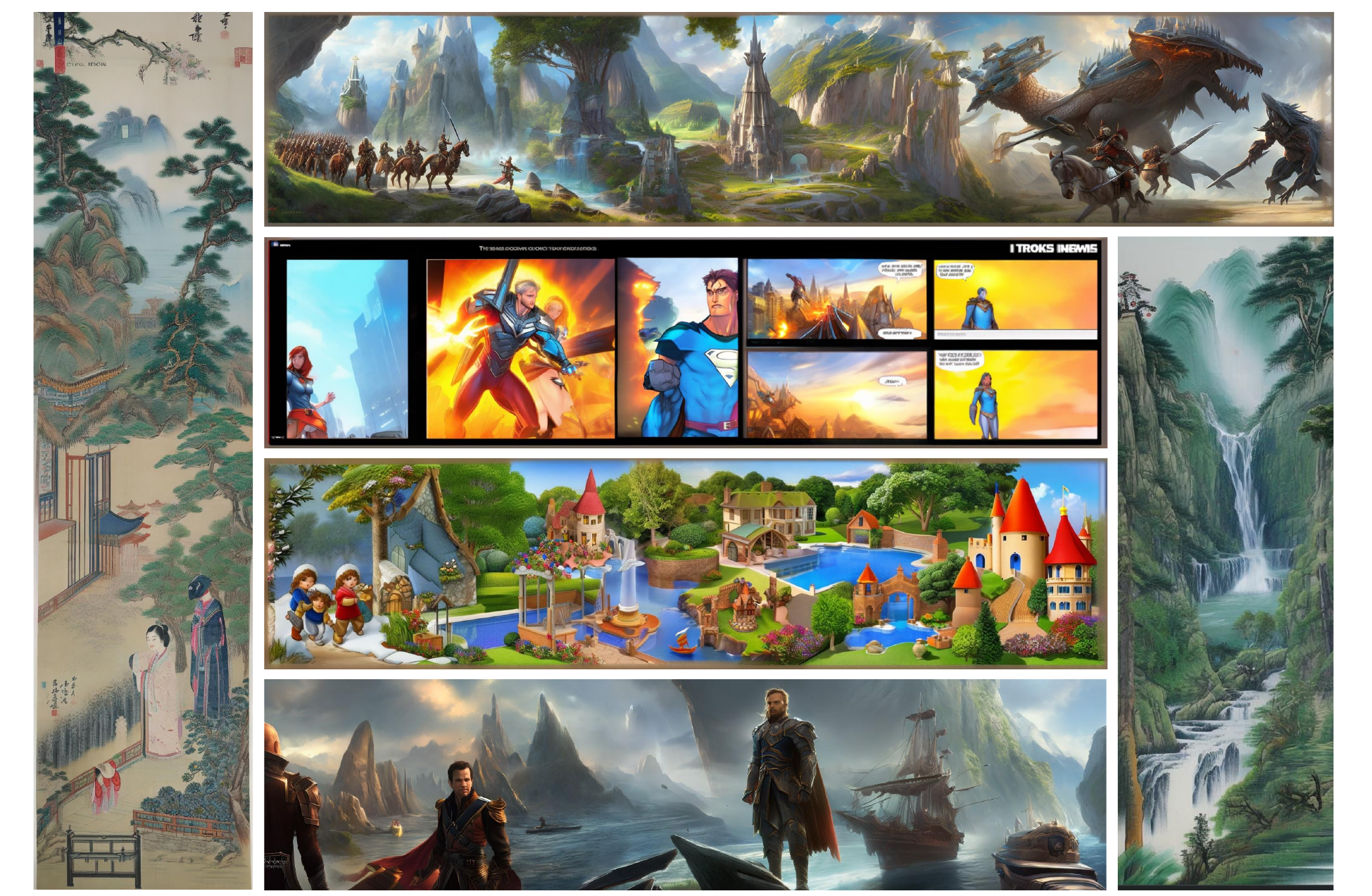}
    \caption{\textbf{\textit{More example results generated by MagicScroll.}} By providing control over style, concept, and layout at all foreground, midground, and background levels, our framework can meet the needs of visual storytelling content generation in various scenarios. Full text prompts can be found in the supplementary material.}
    \label{fig:gallery}
\end{figure*}



\noindent\textbf{Strengths.}
Our framework exemplifies several core advantages corresponding to the key facets of visual storytelling, including content richness, controllability, and coherence. Through a novel semantic-aware denoising process, MagicScroll can unfold a story into a nontypical aspect-ratio image, preserving its key scenes and objects, featuring rich content and smooth transitions. To leverage the narrative efficacy of images, we also explored the integration of different conditions during the generation process, including various aspects (style, concept, layout), different input modalities (reference image, text, mask), and multiple layers (foreground, midground, background). MagicScroll is a step forward toward controllable generation and panorama synthesis, paving the way for more generalizable frameworks.

\noindent\textbf{Limitations.}
Despite the strengths, there are still some limitations and possible future directions to explore. For example, in processing stories, we may apply tokenizers and encoders better suited for ultra-long texts~\cite{chen2023longlora}. During the generation process, we may exert more conditional controls at various stages or include some pre-trained extra modules~\cite{ham2023modulating, mou2023t2i}. We can also learn a mapping between image features and our parameter space, which may further enhance controllability over visual effects.

\section{Conclusion}
\label{sec:conclusion}

In this paper, we propose MagicScroll, a framework for generating controllable, coherent, engaging nontypical aspect-ratio images for visual storytelling. The pipeline is designed to convert a story to an image with a coherent plot unfolding, focusing on critical scenes and objects. We combine various methods to exert control from different aspects, in multiple layers, conditioned on various modalities. With a clear definition of the task, we provided a benchmark and new metrics tailored for evaluating the quality of generated nontypical aspect-ratio images for storytelling. Through qualitative and quantitative evaluation, we show that MagicScroll outperforms alternative methods, achieving high quality similar to ground truth, as well as visual coherence and engagement with the audience.


\noindent\textbf{Societal Impact.}
Our endeavor not only advances the frontier of controllable generation and visual storytelling but also contributes to a broader discussion stemming from production requirements and industrial demands. We look forward to further generalizing the task and framework and expanding the benchmark and metrics for different visual storytelling scenarios, so we can enhance the synergy between creative practitioners and AI researchers.

{
    \small
    \bibliographystyle{ieeenat_fullname}
    \bibliography{main}

\begin{thebibliography}{55}
\providecommand{\natexlab}[1]{#1}
\providecommand{\url}[1]{\texttt{#1}}
\expandafter\ifx\csname urlstyle\endcsname\relax
  \providecommand{\doi}[1]{doi: #1}\else
  \providecommand{\doi}{doi: \begingroup \urlstyle{rm}\Url}\fi

\bibitem[Ao et~al.(2023)Ao, Zhang, and Liu]{ao2023gesturediffuclip}
Tenglong Ao, Zeyi Zhang, and Libin Liu.
\newblock {GestureDiffuCLIP: Gesture Diffusion Model with CLIP Latents}.
\newblock \emph{arXiv preprint arXiv:2303.14613}, 2023.

\bibitem[Bar-Tal et~al.(2023)Bar-Tal, Yariv, Lipman, and
  Dekel]{bar2023multidiffusion}
Omer Bar-Tal, Lior Yariv, Yaron Lipman, and Tali Dekel.
\newblock {MultiDiffusion: Fusing Diffusion Paths for Controlled Image
  Generation}.
\newblock \emph{arXiv preprint arXiv:2302.08113}, 2023.

\bibitem[Chefer et~al.(2023)Chefer, Alaluf, Vinker, Wolf, and
  Cohen-Or]{chefer2023attend}
Hila Chefer, Yuval Alaluf, Yael Vinker, Lior Wolf, and Daniel Cohen-Or.
\newblock {Attend-and-Excite: Attention-Based Semantic Guidance for
  Text-to-Image Diffusion Models}.
\newblock \emph{ACM Trans. Graph.}, 42\penalty0 (4):\penalty0 1--10, 2023.

\bibitem[Chen et~al.(2023)Chen, Qian, Tang, Lai, Liu, Han, and
  Jia]{chen2023longlora}
Yukang Chen, Shengju Qian, Haotian Tang, Xin Lai, Zhijian Liu, Song Han, and
  Jiaya Jia.
\newblock {LongLoRA: Efficient Fine-tuning of Long-Context Large Language
  Models}, 2023.

\bibitem[Choi et~al.(2021)Choi, Kim, Jeong, Gwon, and Yoon]{choi2021ilvr}
Jooyoung Choi, Sungwon Kim, Yonghyun Jeong, Youngjune Gwon, and Sungroh Yoon.
\newblock {ILVR: Conditioning Method for Denoising Diffusion Probabilistic
  Models}.
\newblock \emph{arXiv preprint arXiv:2108.02938}, 2021.

\bibitem[Couairon et~al.(2023)Couairon, Careil, Cord, Lathuili{\`e}re, and
  Verbeek]{couairon2023zero}
Guillaume Couairon, Marl{\`e}ne Careil, Matthieu Cord, St{\'e}phane
  Lathuili{\`e}re, and Jakob Verbeek.
\newblock {Zero-Shot Spatial Layout Conditioning for Text-to-Image Diffusion
  Models}.
\newblock In \emph{Int. Conf. Comput. Vis.}, pages 2174--2183, 2023.

\bibitem[de~La~Rochelle(2023)]{ebdtheque}
Université de La~Rochelle.
\newblock {e-BDthèque}.
\newblock \url{https://ebdtheque.univ-lr.fr/}, 2023.
\newblock Accessed: November 2, 2023.

\bibitem[Deng et~al.(2023)Deng, Jiang, Qi, Yan, Zhou, Guibas, Anguelov,
  et~al.]{deng2023nerdi}
Congyue Deng, Chiyu Jiang, Charles~R Qi, Xinchen Yan, Yin Zhou, Leonidas
  Guibas, Dragomir Anguelov, et~al.
\newblock {NeRDi: Single-View NeRF Synthesis with Language-Guided Diffusion as
  General Image Priors}.
\newblock In \emph{IEEE Conf. Comput. Vis. Pattern Recog.}, pages 20637--20647,
  2023.

\bibitem[Dhariwal and Nichol(2021)]{dhariwal2021diffusion}
Prafulla Dhariwal and Alexander Nichol.
\newblock {Diffusion Models Beat GANs on Image Synthesis}.
\newblock \emph{Adv. Neural Inform. Process. Syst.}, 34:\penalty0 8780--8794,
  2021.

\bibitem[Gal et~al.(2022)Gal, Alaluf, Atzmon, Patashnik, Bermano, Chechik, and
  Cohen-Or]{gal2022image}
Rinon Gal, Yuval Alaluf, Yuval Atzmon, Or Patashnik, Amit~H Bermano, Gal
  Chechik, and Daniel Cohen-Or.
\newblock {An Image Is Worth One Word: Personalizing Text-to-Image Generation
  Using Textual Inversion}.
\newblock \emph{arXiv preprint arXiv:2208.01618}, 2022.

\bibitem[Ham et~al.(2023)Ham, Hays, Lu, Singh, Zhang, and
  Hinz]{ham2023modulating}
Cusuh Ham, James Hays, Jingwan Lu, Krishna~Kumar Singh, Zhifei Zhang, and
  Tobias Hinz.
\newblock {Modulating Pretrained Diffusion Models for Multimodal Image
  Synthesis}.
\newblock \emph{arXiv preprint arXiv:2302.12764}, 2023.

\bibitem[He et~al.(2023)He, Yang, Chen, Cun, Xia, Zhang, Wang, He, Chen, and
  Shan]{he2023scalecrafter}
Yingqing He, Shaoshu Yang, Haoxin Chen, Xiaodong Cun, Menghan Xia, Yong Zhang,
  Xintao Wang, Ran He, Qifeng Chen, and Ying Shan.
\newblock {ScaleCrafter: Tuning-Free Higher-Resolution Visual Generation with
  Diffusion Models}.
\newblock \emph{arXiv preprint arXiv:2310.07702}, 2023.

\bibitem[Hertz et~al.(2022)Hertz, Mokady, Tenenbaum, Aberman, Pritch, and
  Cohen-Or]{hertz2022prompt}
Amir Hertz, Ron Mokady, Jay Tenenbaum, Kfir Aberman, Yael Pritch, and Daniel
  Cohen-Or.
\newblock {Prompt-to-Prompt Image Editing with Cross Attention Control}.
\newblock \emph{arXiv preprint arXiv:2208.01626}, 2022.

\bibitem[Ho et~al.(2020)Ho, Jain, and Abbeel]{ho2020denoising}
Jonathan Ho, Ajay Jain, and Pieter Abbeel.
\newblock {Denoising Diffusion Probabilistic Models}.
\newblock \emph{Adv. Neural Inform. Process. Syst.}, 33:\penalty0 6840--6851,
  2020.

\bibitem[Hu et~al.(2021)Hu, Shen, Wallis, Allen-Zhu, Li, Wang, Wang, and
  Chen]{hu2021lora}
Edward~J Hu, Yelong Shen, Phillip Wallis, Zeyuan Allen-Zhu, Yuanzhi Li, Shean
  Wang, Lu Wang, and Weizhu Chen.
\newblock {LoRA: Low-Rank Adaptation of Large Language Models}.
\newblock \emph{arXiv preprint arXiv:2106.09685}, 2021.

\bibitem[Inoue et~al.(2023)Inoue, Kikuchi, Simo-Serra, Otani, and
  Yamaguchi]{inoue2023layoutdm}
Naoto Inoue, Kotaro Kikuchi, Edgar Simo-Serra, Mayu Otani, and Kota Yamaguchi.
\newblock {LayoutDM: Discrete Diffusion Model for Controllable Layout
  Generation}.
\newblock In \emph{IEEE Conf. Comput. Vis. Pattern Recog.}, pages 10167--10176,
  2023.

\bibitem[Kaggle(2023)]{movie_scripts_corpus}
Kaggle.
\newblock {Movie Scripts Corpus}.
\newblock \url{https://www.kaggle.com/datasets/gufukuro/movie-scripts-corpus},
  2023.
\newblock Accessed: November 2, 2023.

\bibitem[Kawar et~al.(2023)Kawar, Zada, Lang, Tov, Chang, Dekel, Mosseri, and
  Irani]{kawar2023imagic}
Bahjat Kawar, Shiran Zada, Oran Lang, Omer Tov, Huiwen Chang, Tali Dekel, Inbar
  Mosseri, and Michal Irani.
\newblock {Imagic: Text-Based Real Image Editing with Diffusion Models}.
\newblock In \emph{IEEE Conf. Comput. Vis. Pattern Recog.}, pages 6007--6017,
  2023.

\bibitem[Kingma and Welling(2013)]{kingma2013auto}
Diederik~P Kingma and Max Welling.
\newblock {Auto-Encoding Variational Bayes}.
\newblock \emph{arXiv preprint arXiv:1312.6114}, 2013.

\bibitem[Li and Bansal(2023)]{li2023panogen}
Jialu Li and Mohit Bansal.
\newblock {PanoGen: Text-Conditioned Panoramic Environment Generation for
  Vision-and-Language Navigation}.
\newblock \emph{arXiv preprint arXiv:2305.19195}, 2023.

\bibitem[Li et~al.(2019)Li, Gan, Shen, Liu, Cheng, Wu, Carin, Carlson, and
  Gao]{StoryGAN}
Yitong Li, Zhe Gan, Yelong Shen, Jingjing Liu, Yu Cheng, Yuexin Wu, Lawrence
  Carin, David Carlson, and Jianfeng Gao.
\newblock {StoryGAN: A Sequential Conditional GAN for Story Visualization}.
\newblock In \emph{IEEE Conf. Comput. Vis. Pattern Recog.}, pages 6322--6331,
  2019.

\bibitem[Li et~al.(2023)Li, Liu, Wu, Mu, Yang, Gao, Li, and Lee]{li2023gligen}
Yuheng Li, Haotian Liu, Qingyang Wu, Fangzhou Mu, Jianwei Yang, Jianfeng Gao,
  Chunyuan Li, and Yong~Jae Lee.
\newblock {Gligen: Open-Set Grounded Text-to-Image Generation}.
\newblock In \emph{IEEE Conf. Comput. Vis. Pattern Recog.}, pages 22511--22521,
  2023.

\bibitem[Liu et~al.(2023)Liu, Li, Wu, and Lee]{liu2023visual}
Haotian Liu, Chunyuan Li, Qingyang Wu, and Yong~Jae Lee.
\newblock {Visual Instruction Tuning}.
\newblock \emph{arXiv preprint arXiv:2304.08485}, 2023.

\bibitem[Liu et~al.(2022)Liu, Li, Du, Torralba, and
  Tenenbaum]{liu2022compositional}
Nan Liu, Shuang Li, Yilun Du, Antonio Torralba, and Joshua~B Tenenbaum.
\newblock {Compositional Visual Generation with Composable Diffusion Models}.
\newblock In \emph{Eur. Conf. Comput. Vis.}, pages 423--439. Springer, 2022.

\bibitem[Ma et~al.(2023{\natexlab{a}})Ma, Lewis, Kleijn, and
  Leung]{ma2023directed}
Wan-Duo~Kurt Ma, JP Lewis, W~Bastiaan Kleijn, and Thomas Leung.
\newblock {Directed Diffusion: Direct Control of Object Placement through
  Attention Guidance}.
\newblock \emph{arXiv preprint arXiv:2302.13153}, 2023{\natexlab{a}}.

\bibitem[Ma et~al.(2022{\natexlab{a}})Ma, Wang, Wu, Lyu, Chen, Li, and
  Qiao]{ma2022visual}
Yue Ma, Yali Wang, Yue Wu, Ziyu Lyu, Siran Chen, Xiu Li, and Yu Qiao.
\newblock Visual knowledge graph for human action reasoning in videos.
\newblock In \emph{Proceedings of the 30th ACM International Conference on
  Multimedia}, pages 4132--4141, 2022{\natexlab{a}}.

\bibitem[Ma et~al.(2022{\natexlab{b}})Ma, Yang, Shan, and Li]{ma2022simvtp}
Yue Ma, Tianyu Yang, Yin Shan, and Xiu Li.
\newblock Simvtp: Simple video text pre-training with masked autoencoders.
\newblock \emph{arXiv preprint arXiv:2212.03490}, 2022{\natexlab{b}}.

\bibitem[Ma et~al.(2023{\natexlab{b}})Ma, Cun, He, Qi, Wang, Shan, Li, and
  Chen]{ma2023magicstick}
Yue Ma, Xiaodong Cun, Yingqing He, Chenyang Qi, Xintao Wang, Ying Shan, Xiu Li,
  and Qifeng Chen.
\newblock Magicstick: Controllable video editing via control handle
  transformations.
\newblock \emph{arXiv preprint arXiv:2312.03047}, 2023{\natexlab{b}}.

\bibitem[Ma et~al.(2023{\natexlab{c}})Ma, He, Cun, Wang, Shan, Li, and
  Chen]{ma2023follow}
Yue Ma, Yingqing He, Xiaodong Cun, Xintao Wang, Ying Shan, Xiu Li, and Qifeng
  Chen.
\newblock Follow your pose: Pose-guided text-to-video generation using
  pose-free videos.
\newblock \emph{arXiv preprint arXiv:2304.01186}, 2023{\natexlab{c}}.

\bibitem[Maharana et~al.(2022)Maharana, Hannan, and
  Bansal]{maharana2022storydall}
Adyasha Maharana, Darryl Hannan, and Mohit Bansal.
\newblock {StoryDALL-E: Adapting Pretrained Text-to-Image Transformers for
  Story Continuation}.
\newblock In \emph{Eur. Conf. Comput. Vis.}, pages 70--87. Springer, 2022.

\bibitem[Meng et~al.(2021)Meng, He, Song, Song, Wu, Zhu, and
  Ermon]{meng2021sdedit}
Chenlin Meng, Yutong He, Yang Song, Jiaming Song, Jiajun Wu, Jun-Yan Zhu, and
  Stefano Ermon.
\newblock {SDEdit: Guided Image Synthesis and Editing with Stochastic
  Differential Equations}.
\newblock \emph{arXiv preprint arXiv:2108.01073}, 2021.

\bibitem[Mou et~al.(2023)Mou, Wang, Xie, Zhang, Qi, Shan, and Qie]{mou2023t2i}
Chong Mou, Xintao Wang, Liangbin Xie, Jian Zhang, Zhongang Qi, Ying Shan, and
  Xiaohu Qie.
\newblock {T2I-Adapter: Learning Adapters to Dig out More Controllable Ability
  for Text-to-Image Diffusion Models}.
\newblock \emph{arXiv preprint arXiv:2302.08453}, 2023.

\bibitem[Museum(2023)]{minghuaji}
The~Palace Museum.
\newblock {Minghua Ji}.
\newblock \url{https://minghuaji.dpm.org.cn/}, 2023.
\newblock Accessed: November 2, 2023.

\bibitem[OpenAI(2023)]{openai_model_index}
OpenAI.
\newblock {Model Index for Researchers}.
\newblock \url{https://platform.openai.com/docs/model-index-for-researchers},
  2023.
\newblock Accessed: November 2, 2023.

\bibitem[Pan et~al.(2023)Pan, Tewari, Leimk{\"u}hler, Liu, Meka, and
  Theobalt]{pan2023drag}
Xingang Pan, Ayush Tewari, Thomas Leimk{\"u}hler, Lingjie Liu, Abhimitra Meka,
  and Christian Theobalt.
\newblock {Drag Your GAN: Interactive Point-Based Manipulation on the
  Generative Image Manifold}.
\newblock In \emph{ACM SIGGRAPH 2023 Conference Proceedings}, pages 1--11,
  2023.

\bibitem[Qu et~al.(2023)Qu, Wu, Fei, Nie, and Chua]{qu2023layoutllm}
Leigang Qu, Shengqiong Wu, Hao Fei, Liqiang Nie, and Tat-Seng Chua.
\newblock {LayoutLLM-T2I: Eliciting Layout Guidance from LLM for Text-to-Image
  Generation}.
\newblock \emph{arXiv preprint arXiv:2308.05095}, 2023.

\bibitem[Radford et~al.(2021)Radford, Kim, Hallacy, Ramesh, Goh, Agarwal,
  Sastry, Askell, Mishkin, Clark, et~al.]{radford2021learning}
Alec Radford, Jong~Wook Kim, Chris Hallacy, Aditya Ramesh, Gabriel Goh,
  Sandhini Agarwal, Girish Sastry, Amanda Askell, Pamela Mishkin, Jack Clark,
  et~al.
\newblock {Learning Transferable Visual Models from Natural Language
  Supervision}.
\newblock In \emph{International conference on machine learning}, pages
  8748--8763. PMLR, 2021.

\bibitem[Rahman et~al.(2023)Rahman, Lee, Ren, Tulyakov, Mahajan, and
  Sigal]{makeastory}
Tanzila Rahman, Hsin-Ying Lee, Jian Ren, Sergey Tulyakov, Shweta Mahajan, and
  Leonid Sigal.
\newblock {Make-A-Story: Visual Memory Conditioned Consistent Story
  Generation}.
\newblock In \emph{IEEE Conf. Comput. Vis. Pattern Recog.}, pages 2493--2502,
  2023.

\bibitem[Rombach et~al.(2022)Rombach, Blattmann, Lorenz, Esser, and
  Ommer]{rombach2022high}
Robin Rombach, Andreas Blattmann, Dominik Lorenz, Patrick Esser, and Bj{\"o}rn
  Ommer.
\newblock {High-Resolution Image Synthesis with Latent Diffusion Models}.
\newblock In \emph{IEEE Conf. Comput. Vis. Pattern Recog.}, pages 10684--10695,
  2022.

\bibitem[Ruiz et~al.(2023)Ruiz, Li, Jampani, Pritch, Rubinstein, and
  Aberman]{ruiz2023dreambooth}
Nataniel Ruiz, Yuanzhen Li, Varun Jampani, Yael Pritch, Michael Rubinstein, and
  Kfir Aberman.
\newblock {Dreambooth: Fine Tuning Text-to-Image Diffusion Models for
  Subject-Driven Generation}.
\newblock In \emph{IEEE Conf. Comput. Vis. Pattern Recog.}, pages 22500--22510,
  2023.

\bibitem[Singer et~al.(2022)Singer, Polyak, Hayes, Yin, An, Zhang, Hu, Yang,
  Ashual, Gafni, et~al.]{singer2022make}
Uriel Singer, Adam Polyak, Thomas Hayes, Xi Yin, Jie An, Songyang Zhang, Qiyuan
  Hu, Harry Yang, Oron Ashual, Oran Gafni, et~al.
\newblock {Make-A-Video: Text-to-Video Generation without Text-Video Data}.
\newblock \emph{arXiv preprint arXiv:2209.14792}, 2022.

\bibitem[Sohn et~al.(2023)Sohn, Ruiz, Lee, Chin, Blok, Chang, Barber, Jiang,
  Entis, Li, et~al.]{sohn2023styledrop}
Kihyuk Sohn, Nataniel Ruiz, Kimin Lee, Daniel~Castro Chin, Irina Blok, Huiwen
  Chang, Jarred Barber, Lu Jiang, Glenn Entis, Yuanzhen Li, et~al.
\newblock {StyleDrop: Text-to-Image Generation in Any Style}.
\newblock \emph{arXiv preprint arXiv:2306.00983}, 2023.

\bibitem[Song et~al.(2020{\natexlab{a}})Song, Meng, and
  Ermon]{song2020denoising}
Jiaming Song, Chenlin Meng, and Stefano Ermon.
\newblock {Denoising Diffusion Implicit Models}.
\newblock \emph{arXiv preprint arXiv:2010.02502}, 2020{\natexlab{a}}.

\bibitem[Song et~al.(2020{\natexlab{b}})Song, Rui~Tam, Chen, Lu, and Shuai]{sv}
Yun-Zhu Song, Zhi Rui~Tam, Hung-Jen Chen, Huiao-Han Lu, and Hong-Han Shuai.
\newblock {Character-Preserving Coherent Story Visualization}.
\newblock In \emph{Eur. Conf. Comput. Vis.}, page 18–33, Berlin, Heidelberg,
  2020{\natexlab{b}}. Springer-Verlag.

\bibitem[Stewart(2023)]{compel}
Damian Stewart.
\newblock {Compel}.
\newblock \url{https://github.com/damian0815/compel}, 2023.
\newblock GitHub repository, Accessed: November 2, 2023.

\bibitem[Wang et~al.(2022{\natexlab{a}})Wang, Yang, Men, Lin, Bai, Li, Ma,
  Zhou, Zhou, and Yang]{wang2022ofa}
Peng Wang, An Yang, Rui Men, Junyang Lin, Shuai Bai, Zhikang Li, Jianxin Ma,
  Chang Zhou, Jingren Zhou, and Hongxia Yang.
\newblock {OFA: Unifying Architectures, Tasks, and Modalities Through a Simple
  Sequence-to-Sequence Learning Framework}.
\newblock In \emph{International Conference on Machine Learning}, pages
  23318--23340. PMLR, 2022{\natexlab{a}}.

\bibitem[Wang et~al.(2022{\natexlab{b}})Wang, Li, Salesin, Snavely, Curless,
  and Kontkanen]{wang20223d}
Qianqian Wang, Zhengqi Li, David Salesin, Noah Snavely, Brian Curless, and
  Janne Kontkanen.
\newblock {3D Moments from Near-Duplicate Photos}.
\newblock In \emph{IEEE Conf. Comput. Vis. Pattern Recog.}, pages 3906--3915,
  2022{\natexlab{b}}.

\bibitem[Wang et~al.(2023{\natexlab{a}})Wang, Saharia, Montgomery, Pont-Tuset,
  Noy, Pellegrini, Onoe, Laszlo, Fleet, Soricut, et~al.]{wang2023imagen}
Su Wang, Chitwan Saharia, Ceslee Montgomery, Jordi Pont-Tuset, Shai Noy,
  Stefano Pellegrini, Yasumasa Onoe, Sarah Laszlo, David~J Fleet, Radu Soricut,
  et~al.
\newblock {Imagen Editor and Editbench: Advancing and Evaluating Text-Guided
  Image Inpainting}.
\newblock In \emph{IEEE Conf. Comput. Vis. Pattern Recog.}, pages 18359--18369,
  2023{\natexlab{a}}.

\bibitem[Wang et~al.(2023{\natexlab{b}})Wang, Wang, Xie, Qi, Shan, Wang, and
  Luo]{wang2023styleadapter}
Zhouxia Wang, Xintao Wang, Liangbin Xie, Zhongang Qi, Ying Shan, Wenping Wang,
  and Ping Luo.
\newblock {StyleAdapter: A Single-Pass LoRA-Free Model for Stylized Image
  Generation}.
\newblock \emph{arXiv preprint arXiv:2309.01770}, 2023{\natexlab{b}}.

\bibitem[Weng et~al.(2019)Weng, Curless, and
  Kemelmacher-Shlizerman]{weng2019photo}
Chung-Yi Weng, Brian Curless, and Ira Kemelmacher-Shlizerman.
\newblock {Photo Wake-Up: 3D Character Animation from a Single Photo}.
\newblock In \emph{IEEE Conf. Comput. Vis. Pattern Recog.}, pages 5908--5917,
  2019.

\bibitem[Wu et~al.(2023)Wu, Nakashima, and Garcia]{wu2023not}
Yankun Wu, Yuta Nakashima, and Noa Garcia.
\newblock {Not Only Generative Art: Stable Diffusion for Content-Style
  Disentanglement in Art Analysis}.
\newblock In \emph{Proceedings of the 2023 ACM International Conference on
  Multimedia Retrieval}, pages 199--208, 2023.

\bibitem[Ye et~al.(2023)Ye, Zhang, Liu, Han, and Yang]{ye2023ip}
Hu Ye, Jun Zhang, Sibo Liu, Xiao Han, and Wei Yang.
\newblock {IP-Adapter: Text Compatible Image Prompt Adapter for Text-to-Image
  Diffusion Models}.
\newblock \emph{arXiv preprint arXiv:2308.06721}, 2023.

\bibitem[Zhang et~al.(2023)Zhang, Rao, and Agrawala]{zhang2023adding}
Lvmin Zhang, Anyi Rao, and Maneesh Agrawala.
\newblock {Adding Conditional Control to Text-to-Image Diffusion Models}.
\newblock In \emph{Int. Conf. Comput. Vis.}, pages 3836--3847, 2023.

\bibitem[Zhao et~al.(2023)Zhao, Chen, Chen, Bao, Hao, Yuan, and
  Wong]{zhao2023uni}
Shihao Zhao, Dongdong Chen, Yen-Chun Chen, Jianmin Bao, Shaozhe Hao, Lu Yuan,
  and Kwan-Yee~K Wong.
\newblock {Uni-ControlNet: All-in-One Control to Text-to-Image Diffusion
  Models}.
\newblock \emph{arXiv preprint arXiv:2305.16322}, 2023.

\bibitem[Zheng et~al.(2023)Zheng, Zhou, Li, Qi, Shan, and
  Li]{zheng2023layoutdiffusion}
Guangcong Zheng, Xianpan Zhou, Xuewei Li, Zhongang Qi, Ying Shan, and Xi Li.
\newblock {LayoutDiffusion: Controllable Diffusion Model for Layout-to-Image
  Generation}.
\newblock In \emph{IEEE Conf. Comput. Vis. Pattern Recog.}, pages 22490--22499,
  2023.

\end{thebibliography}
}

\clearpage
\setcounter{page}{1}
\maketitlesupplementary




\begin{figure*}[bp]
\centering
 \begin{minipage}{0.48\linewidth}

 	\centerline{\includegraphics[width=\textwidth]{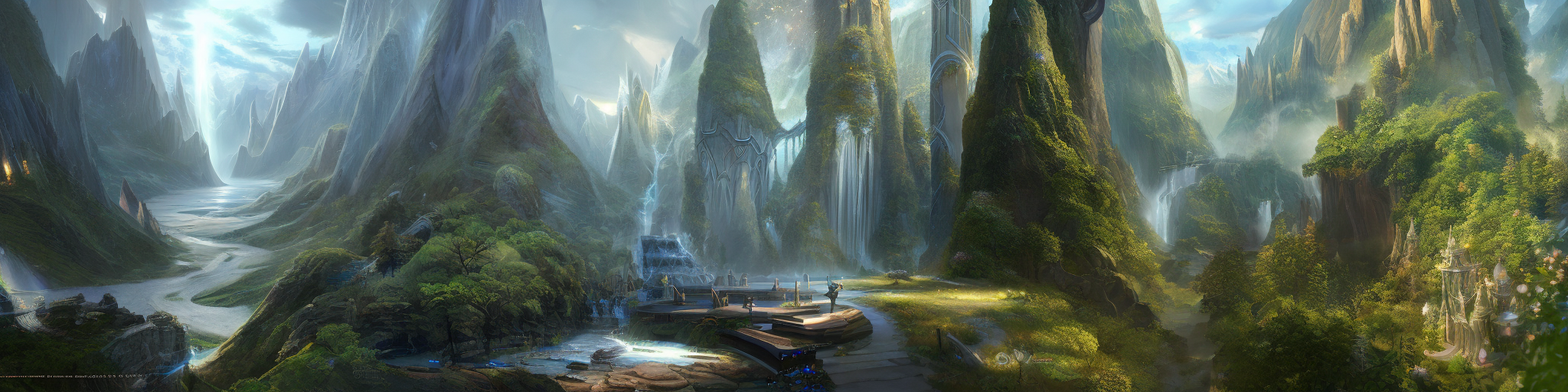}}
 	\vspace{3pt}
   	\centerline{\includegraphics[width=\textwidth]{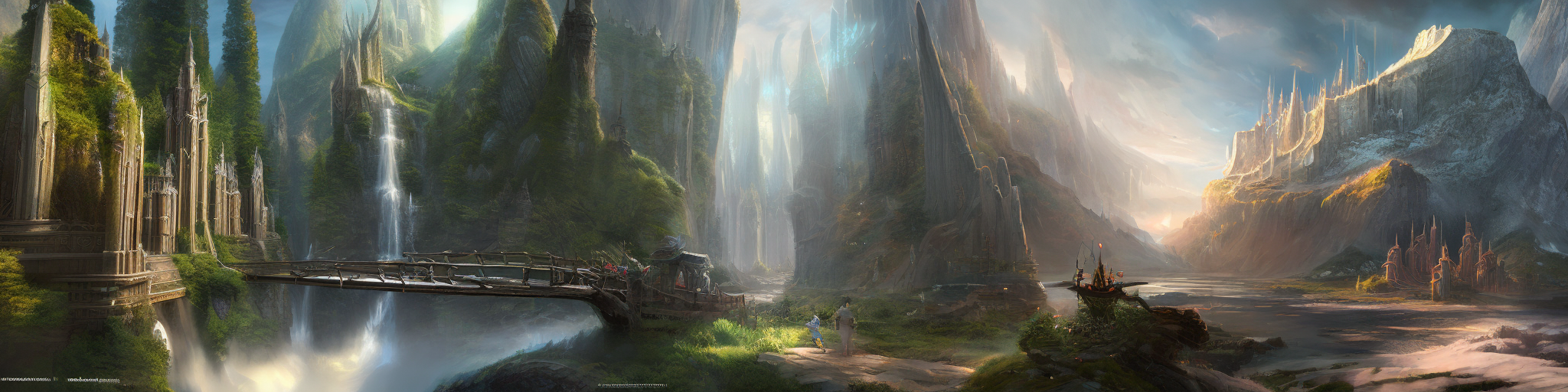}}
 	\vspace{3pt}
   	\centerline{\includegraphics[width=\textwidth]{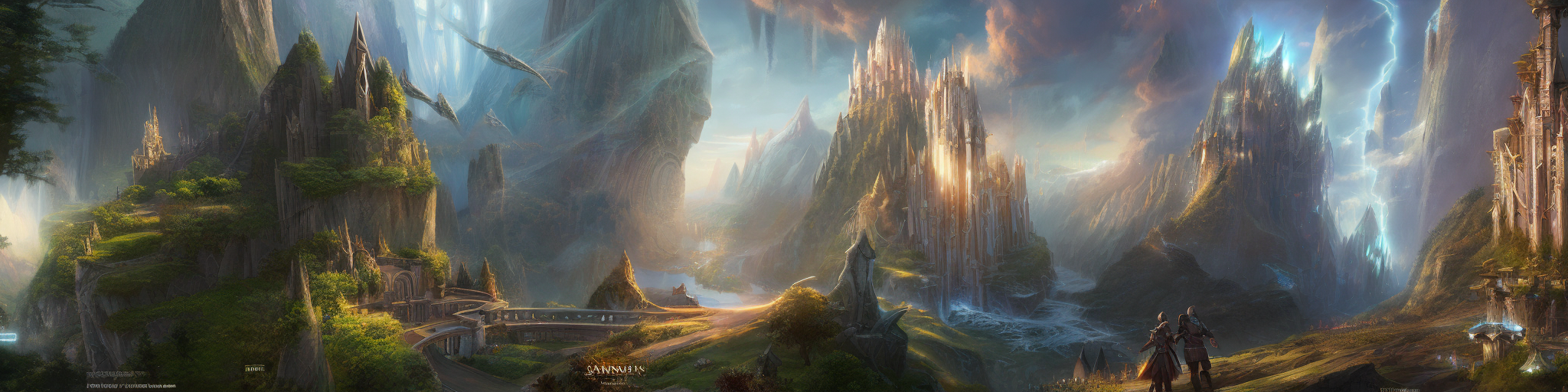}}
 	\vspace{3pt}
 	\centerline{\includegraphics[width=\textwidth]{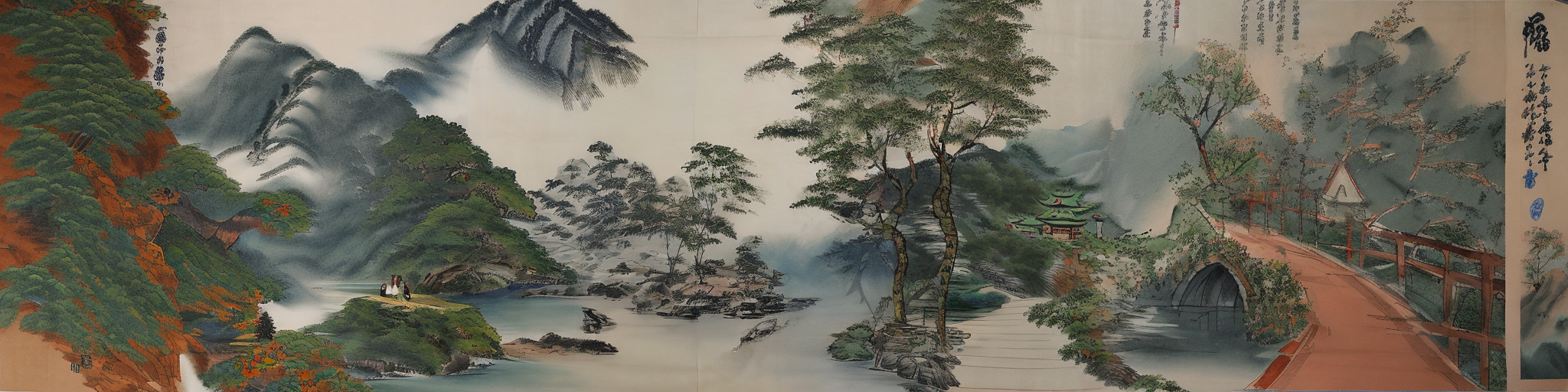}}
 	\vspace{3pt}
   	\centerline{\includegraphics[width=\textwidth]{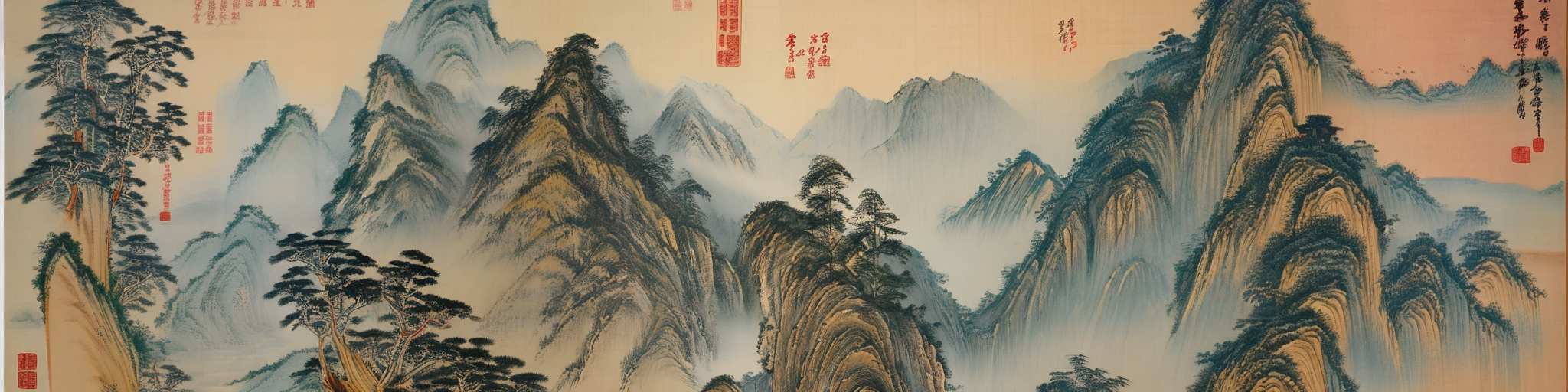}}
    \vspace{3pt}
\centerline{\includegraphics[width=\textwidth]{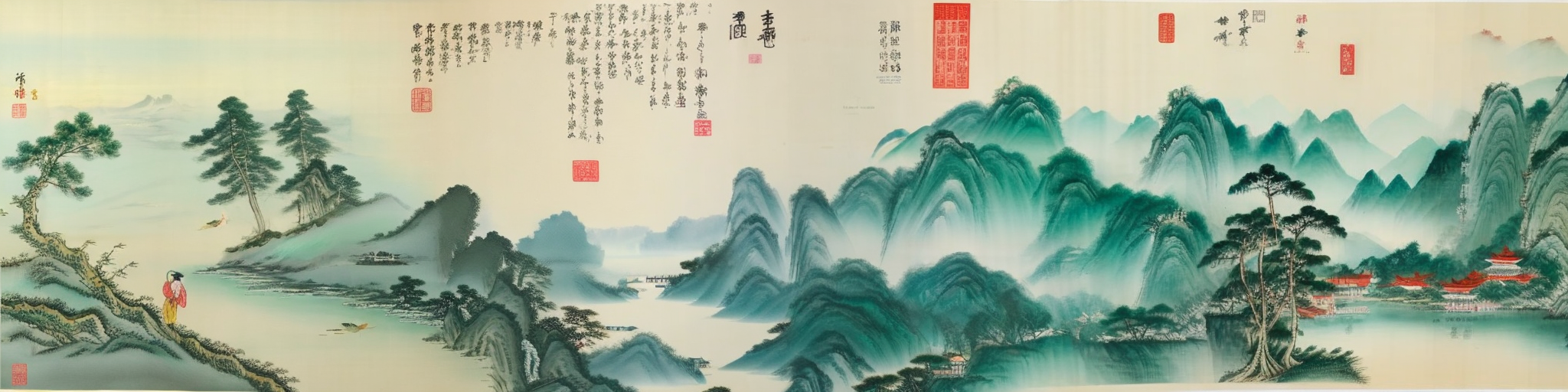}}
 	\vspace{3pt}
 	\centerline{\includegraphics[width=\textwidth]{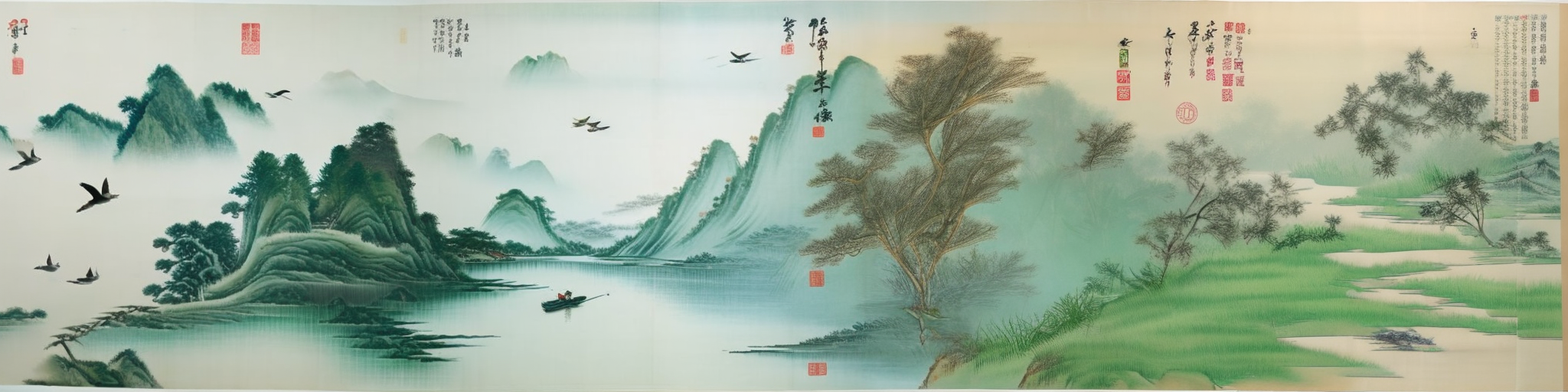}}
 	\vspace{3pt}
 	\centerline{\includegraphics[width=\textwidth]{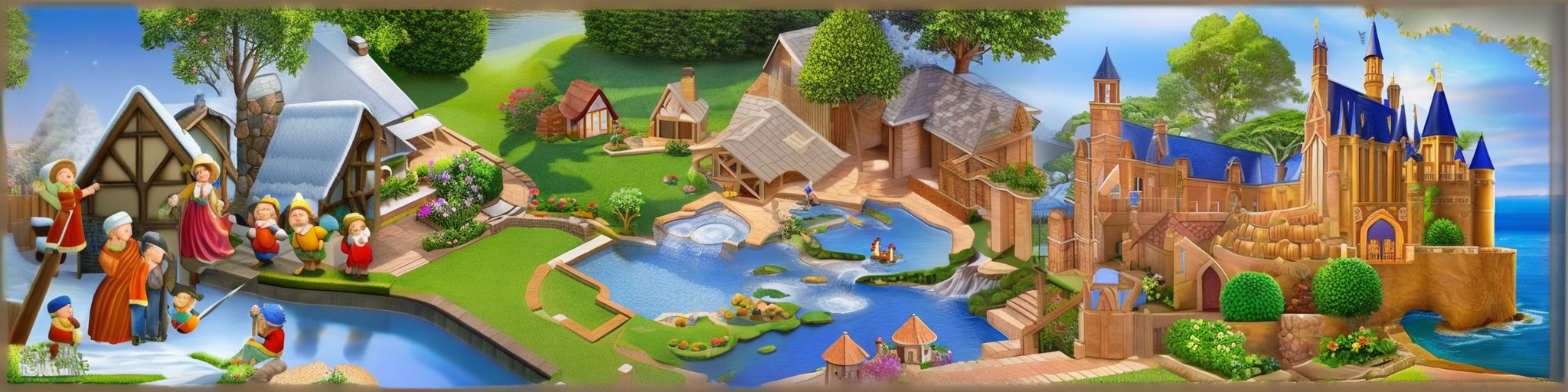}}
\end{minipage}
\begin{minipage}{0.48\linewidth}
 	\centerline{\includegraphics[width=\columnwidth]{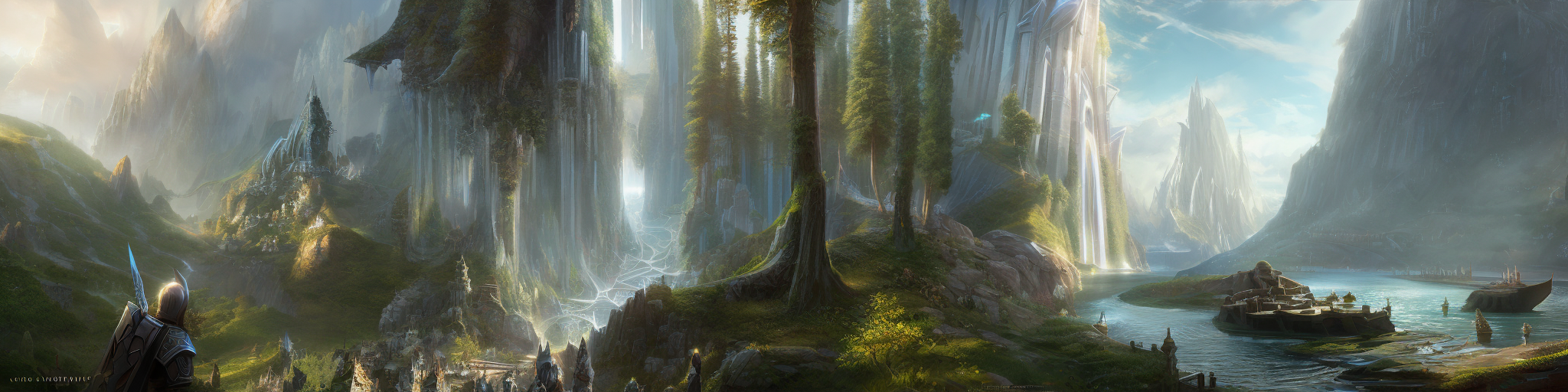}}
 	\vspace{3pt}
 	\centerline{\includegraphics[width=\columnwidth]{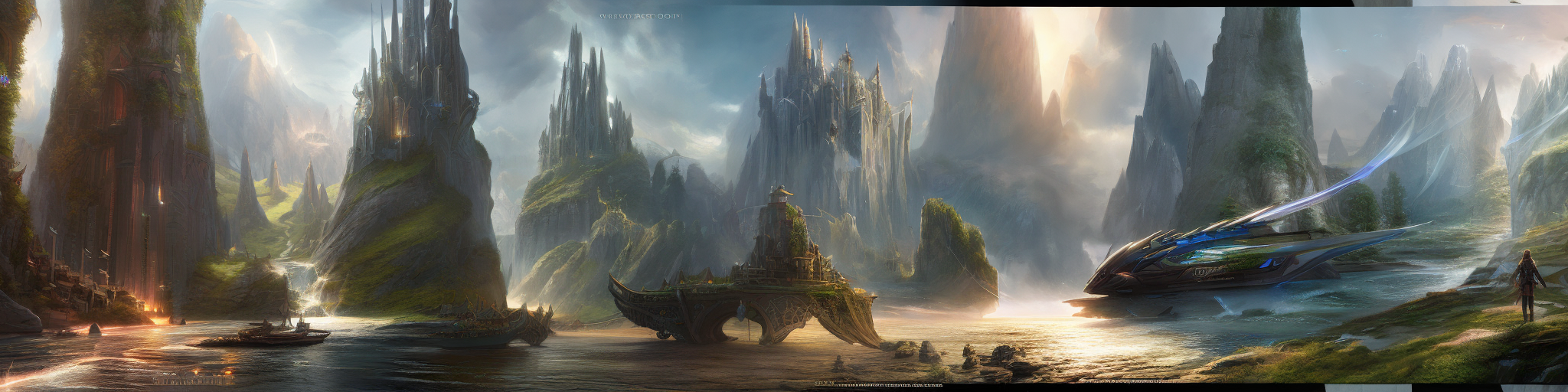}}
 	\vspace{3pt}
 	\centerline{\includegraphics[width=\columnwidth]{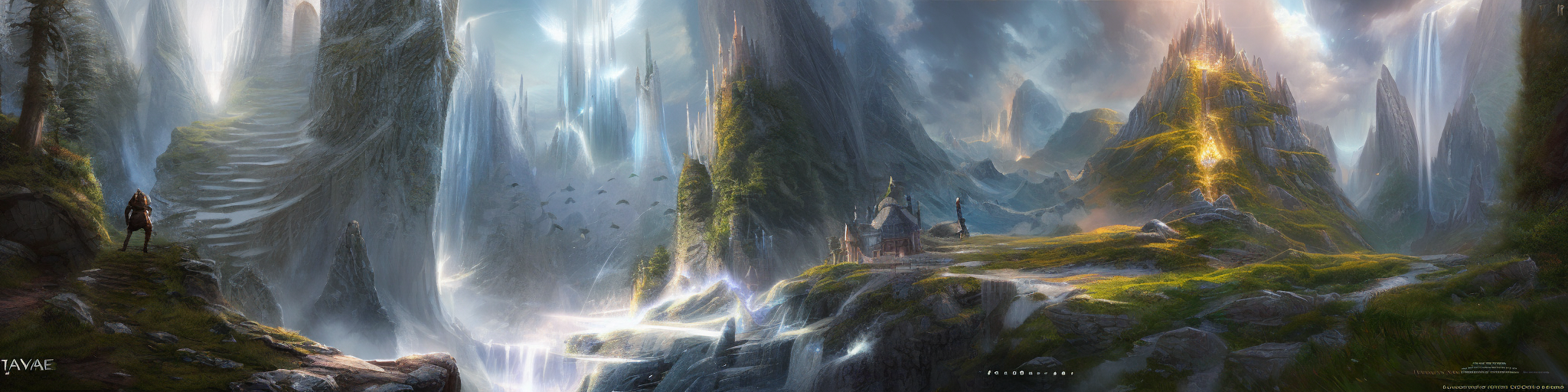}}
 	\vspace{3pt}
   	\centerline{\includegraphics[width=\columnwidth]{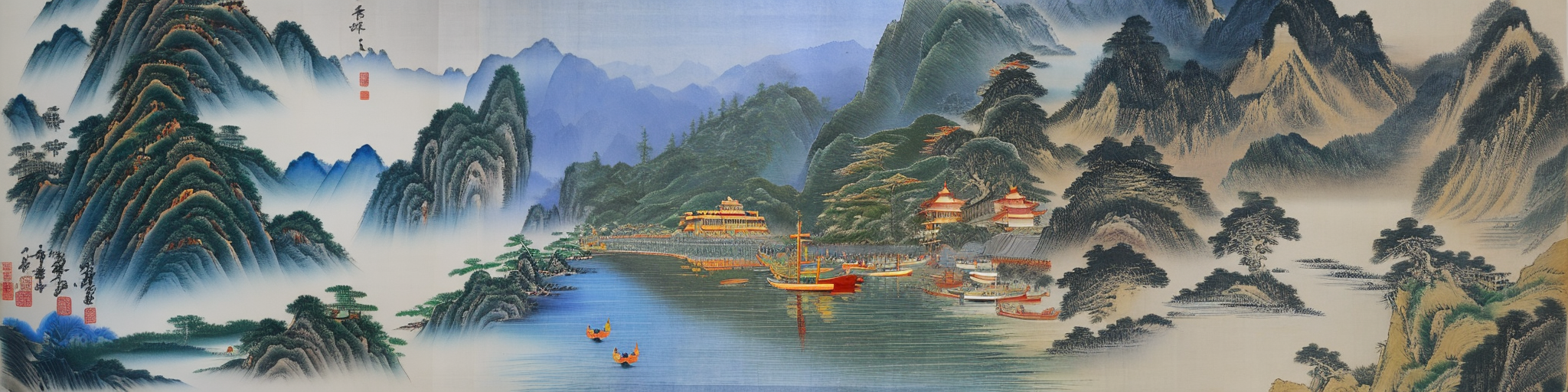}}
 	\vspace{3pt}
 	\centerline{\includegraphics[width=\columnwidth]{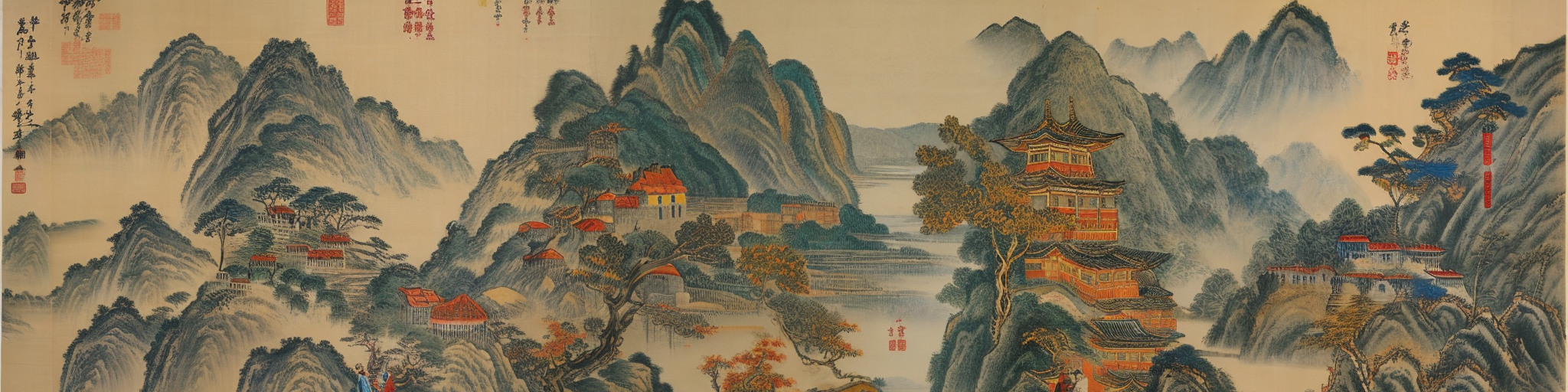}}
 	\vspace{3pt}
 	\centerline{\includegraphics[width=\columnwidth]{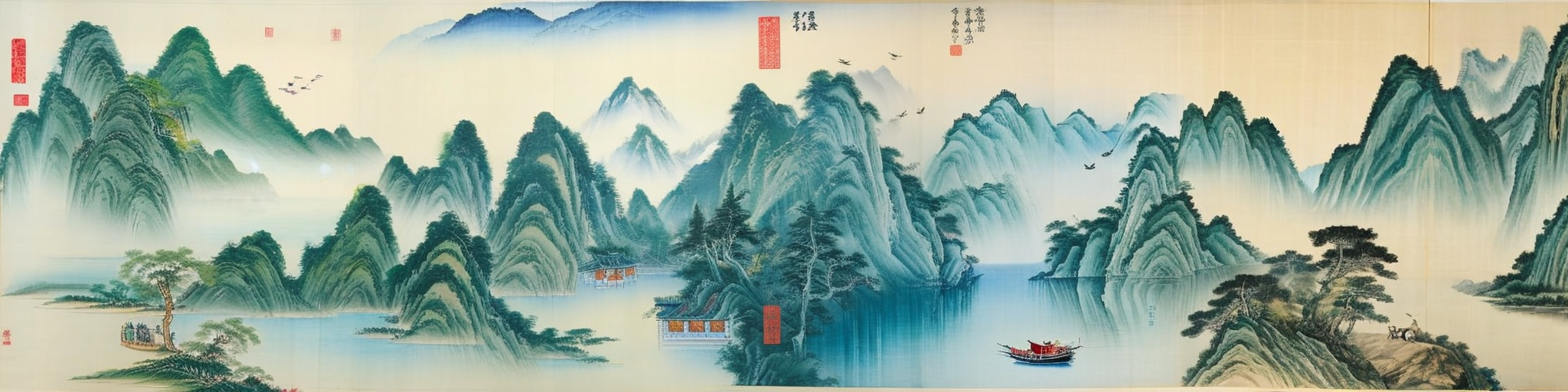}}
 	\vspace{3pt}
 	\centerline{\includegraphics[width=\columnwidth]{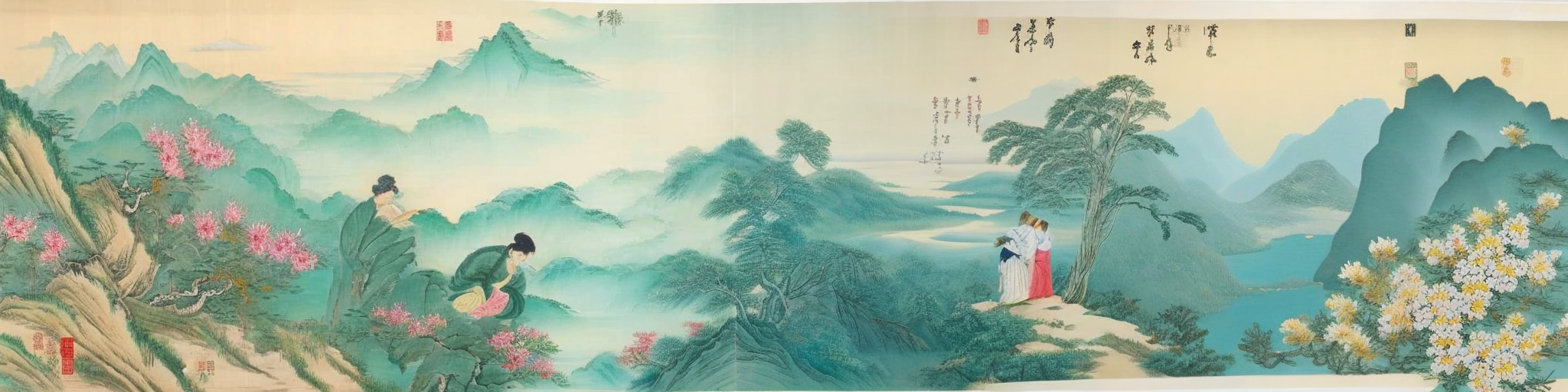}}
 	\vspace{3pt}
   	\centerline{\includegraphics[width=\columnwidth]{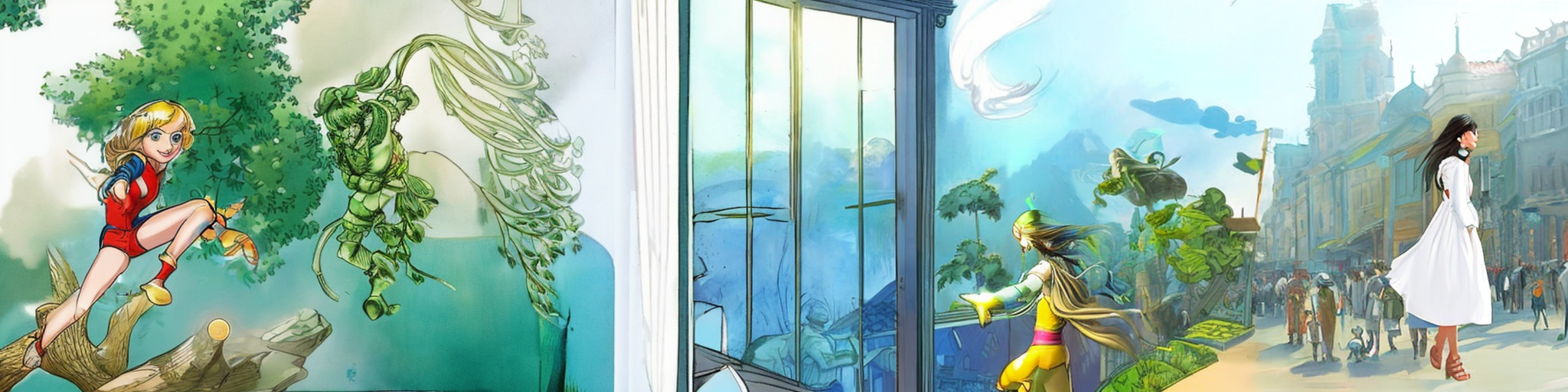}}
\end{minipage}
    \caption{\textbf{\textit{More results generated by MagicScroll.}}}
    \label{fig:more}
\end{figure*}

\twocolumn[{
\textbf{Text prompts for Figure~\ref{fig:gallery}.}
\begin{itemize}
    \item Chinese painting (left): 
In a deep mountain enclave, where pines and cypresses thrive,
Amidst swirling clouds and mist, distant rocks and peaks arrive.
A dwelling stands in solitude, with a simple architectural grace,
Beyond bamboo fences, children play in a joyous embrace.
Beneath the trees, two figures converse with laughter and delight,
Together forming a scene of tranquil seclusion, warm and bright.
    \item Chinese painting (right): 
In deep mountains, a waterfall descends from great height,
Winding its way down, into a crystal pool, pure and bright.
Water droplets dance, a splash of liquid light,
In the distance, evergreen pines, in the foreground, rocks stand upright.
The springwater murmurs, weaving a melody with stones,
Surrounded by bushes, a charming scene it owns.
    \item Cinematic panorama (top): 
Amidst the sprawling castles, a troop of cavalry embarks on a quest. They traverse distant mountains, weaving through dense forests, passing rivers, castles, and perilous peaks. On the other end awaits a colossal dragon and its minions, their forms menacing, teeth bared, claws ready—intent on slaying all who dare to invade.
    \item Cinematic panorama (bottom):
Captain David and First Mate Jess set sail for a distant voyage. Their ship gazes upon far-off peaks, navigating through thick veils of mist, braving treacherous reefs. David and Jess stand boldly at the bow, unfolding a majestic scene like a grand painting coming to life.
    \item Comic strip (upper middle):
Early in the morning, the hero Catherine received a call to save the world. Armed and ready, she joined forces with Jack and headed towards the heart of the city. Gazing at the distant smoke, Jack's expression turned solemn as he thought of the lives lost. Despite the gravity of the situation, they bravely rescued the city, fulfilling their mission. The duo stood victorious, having overcome the challenges in their path.
    \item Comic strip (lower middle):
In a fairy tale, there exists a village of toys, adorned with beautiful castles, lakes, and flowers. People picnic in the fields, row boats on the lake, and celebrate harvest days in the snowy season. In this village, docks, fountains, and flower houses are constructed to embellish the tranquil life of its inhabitants.
\end{itemize}
}]

\end{document}